\documentclass[10pt,twocolumn,letterpaper]{article}

\usepackage{cvpr}              

\usepackage{graphicx}
\usepackage{url}
\usepackage{tcolorbox}
\usepackage{booktabs}
\usepackage{caption}
\usepackage{enumitem}
\usepackage{makecell}
\usepackage{threeparttable}
\usepackage[accsupp]{axessibility} 

\usepackage{xcolor}
\usepackage{soul}
\sethlcolor{yellow} 

\definecolor{cvprblue}{rgb}{0.21,0.49,0.74}
\usepackage[pagebackref,breaklinks,colorlinks,allcolors=cvprblue]{hyperref}

\def\paperID{14267} 
\def\confName{CVPR}
\def\confYear{2026}

\title{Mitigating Batch Effects in Histopathology \\ via Language-Mediated Robust Embedding Generation}

\author{ Yishu Zhang, Shushan Wu, Zhenzhong Zhang, Didong Li, Huaxiu Yao, Yun Li, Iain Carmichael,\\ Katherine A. Hoadley, Hongtu Zhu, 
\setcounter{footnote}{1}
Di Wu\thanks{Corresponding authors}, Daiwei Zhang\footnotemark[2]\vspace{0.4em}\\ 
University of North Carolina at Chapel Hill\\
{\tt\small \{yzhang4@, shushan\_wu@med., zhang3zz@, didongli@, huaxiu@cs.,}\\
{\tt\small yun\_li@med., iain@, hoadley@med., htzhu@email., dwu@, daiwei\_zhang@med.\}unc.edu} }

\begin{document}
\maketitle

\begin{abstract}
Pathology foundation models (PFMs) have demonstrated strong potential across clinical and scientific applications, yet their performance is often hindered by batch effects, which are non-biological variations across tissue source institutions (TSIs) that distort learned feature representations and impair generalization. Conventional mitigation strategies, such as stain normalization, offer limited success in addressing these high-dimensional, complex artifacts. We present GLMP (General-purpose LLM-Mediated Pathology model), a novel framework that generates robust numerical embeddings from histology image patches through an intermediate textual representation. By leveraging pretrained general-purpose multimodal large language models (MLLMs) and text encoders, GLMP effectively prioritizes biologically meaningful signals over TSI-specific artifacts, thereby improving cross-institutional generalization. To our knowledge, GLMP is the first pathology model to use text descriptions of histological features as an intermediate representation for generating numerical embeddings from histology images. Our results highlight the untapped potential of broad-domain, non-specialized MLLMs in computational pathology and introduce a new paradigm for building versatile, generalizable, and robust pathology models.
\end{abstract}

\section{Introduction}

Pathology foundation models (PFMs), pre-trained on large-scale unannotated histology data, have reshaped computational pathology by enabling diverse downstream clinical and scientific tasks. State-of-the-art PFMs are primarily vision encoders trained via self-supervised learning on hematoxylin-and-eosin (H\&E) whole-slide images (WSIs) \citep{chen2024uni, xu2024gigapath, nechaev2024hibou, zimmermann2024virchow2, ma2025generalizable}. Beyond patch-level encoders, PFMs integrate slide-level context through attention-based multiple instance learning or hierarchical transformers that aggregate patch embeddings into whole-slide representations \citep{wang2024pfm, vaidya2025threads, xu2024wholeslide, ding2025multimodal}. Multimodal extensions integrate additional stains \citep{dippel2024rudolfv, hua2024pathoduet}, molecular profiles \citep{xu2024multimodal, vaidya2025threads}, or paired image-text data \citep{lu2024visual, xiang2025vision}. Pairing these encoders with large language models (LLMs) powers generative systems such as PathChat+ \citep{chen2025pathchatplus}, SlideChat \citep{chen2025slidechat}, and PRISM2 \citep{shaikovski2025prism2}. Collectively, these capabilities position PFMs as a transformative tool for applications such as tissue classification, disease subtyping, and survival analysis, with promising potential for supporting clinical decision-making and advancing precision medicine in pathology.

Yet PFMs are susceptible to batch effects, which are non-biological artifacts linked to tissue source institutions (TSIs) \citep{Howard2021, komen2024batch, deJong2025, komen2025towards} that commonly arise from variations in tissue fixation, processing, staining, and scanner hardware \citep{hoque2024stain, chai2025impact}. Recent studies highlight the deleterious impact of these batch effects, showing that TSI signatures in modern PFMs (including their newer iterations) can dominate embedding geometry, create bias in downstream tasks, and harm out-of-distribution performance, even despite traditional stain normalization \citep{lin2025impact, deJong2025, alfasly2025validation}. These artifacts can correlate with disease severity or other clinical variables, making TSI-specific signatures difficult to disentangle from biological signals. Consequently, state-of-the-art PFMs encode TSI information more strongly than biomedical attributes such as disease status \citep{deJong2025, komen2025towards}, leading to inflated accuracy estimates via site-specific shortcuts \citep{Howard2021} and causing real-world failure under clinical distribution shifts. Such failures propagate clinical bias when institutional signatures are associated with clinical variables of interest \citep{Dehkharghanian2023}, and risk unsafe generalization across different labs and devices in high-stake medical settings \citep{lin2025impact, alfasly2025validation}. For example, artifact-dominated embeddings can cause image retrieval systems to return cases based on staining similarity rather than morphology, and can undermine multimodal learning framework that integrates histology images with other tissue modalities (e.g. spatial transcriptomics), where biologically coherent representations across slides is essential \citep{nishimura2025learning,huang2025scalable}. Despite these risks and obstacles caused by batch effects, many studies in computational pathology continue to benchmark PFMs using data from a single source institution, providing limited evaluation of their generalizability across TSIs. Mitigating batch effects and assessing their effects on pathology models is therefore critical for generalizable and safe deployment.

Various techniques have been developed to mitigate batch effects in histology images. These solutions primarily rely on color normalization \citep{reinhard2001color, macenko2009method, vahadane2016structure} and augmentation strategies that broaden model exposure to stain variability \citep{shen2022randstainna, marini2023data}. However, as these pixel-level remedies cannot remove higher-order artifacts introduced by tissue processing and scanner hardware, TSI signatures persist in learned representations of most PFMs \citep{keller2023tissue, komen2024batch}.

\begin{figure*}[t]
  \centering
  \includegraphics[width=\textwidth]{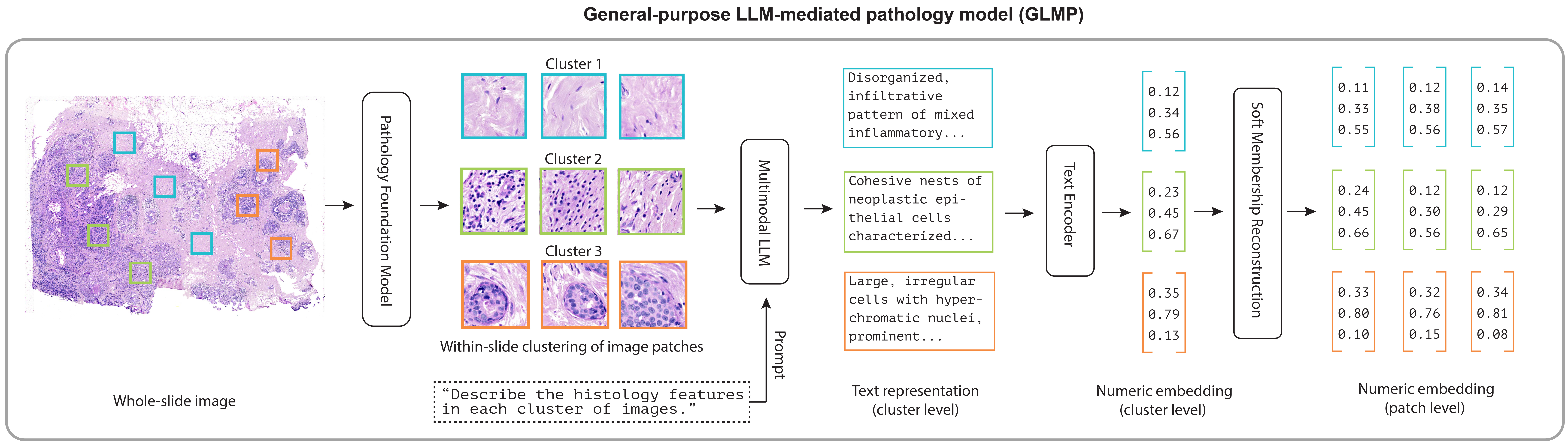}
  \caption{The GLMP framework. Instead of directly encoding a histology image patch, GLMP first uses a pathology foundation model (e.g. Virchow2) to cluster patches in a whole-slide image into histology clusters. An MLLM (e.g. Gemini 2.5 Pro) then generates a text description of each cluster's biological content, which is converted into a cluster-level embedding by a text encoder (e.g. Gemini Embedding). The final, unique embedding for each patch is then computed as a weighted average of these cluster-level embeddings, where the weights are proportional to soft cluster memberships.}
\label{fig:schematic}
\end{figure*}

To overcome the limitations of existing PFMs and batch effect correction methods, we developed the General-purpose LLM-Mediated Pathology model (GLMP) (\autoref{fig:schematic}). In contrast to standard PFMs that map a histology image patch directly to a numeric feature vector, our approach generates an intermediate textual description of given histology image patches using a pretrained general-purpose MLLM (e.g. Gemini 2.5 Pro \cite{comanici2025gemini}). A structured prompt guides the MLLM to describe only biological characteristics, such as cellular morphology and tissue architecture, and filter out non-biological information, such as stain colors and image brightness. The resulting text is then encoded into a final embedding using a pretrained general-purpose text encoder. Rather than directly assessing the clinical accuracy of the intermediate text, we measure the utility of our text-mediated model by the performance of the final embeddings on various downstream tasks. This work demonstrates that compared with existing PFMs, GLMP prioritizes genuine biological differences in tissues over artificial technical artifacts due to TSIs, yielding more robust and consistent feature representations for downstream analysis. 

\section{Related work}

\textbf{Advanced methods for mitigating batch effects in WSIs.} Several advanced strategies integrate artifact mitigation directly into the model's architecture or training process. One such strategy is based on Generative Adversarial Networks (GANs), which learn an unsupervised image-to-image translation to harmonize stain appearance while preserving morphology \citep{zanjani2018stain, shaban2019}. Another approach adapts the Domain-Adversarial Neural Network (DANN) framework \citep{ganin2016dann} for histopathology, leveraging an auxiliary classifier with a gradient reversal layer to produce features that are task-predictive yet invariant to site-specific artifacts \citep{otalora2019staining, lafarge2019learning}. This process requires joint training with domain labels from multiple sites to enforce invariance. Recent work learns semantically invariant features by aligning histology images with paired pathology texts \citep{huang2023visual, ikezogwo2023quilt, lu2024visual}. However, the visual embeddings from these text-alignment models can still retain TSI-specific artifacts \citep{komen2025towards}.

\noindent\textbf{Application of MLLMs for pathology.}
Recent work has explored MLLMs for both direct classification and feature representation in pathology. \citet{ferber2024icl} show that in-context learning enables general-purpose MLLMs such as GPT4 to outperform task-specific PFMs on tile-level tasks, but this approach does not provide task-agnostic numeric embeddings. The MLLM4PUE framework \citep{zhou2025mllm4pue} introduces an information bottleneck by prompting an MLLM to generate a single-word summary of an image and uses the word's hidden state as the embedding, but this method constrains the semantic representation to a single token, which severely limits the richness of the embedded histopathological information, and requires access to the MLLM's hidden state, which is not available for frontier proprietary models. 

\noindent\textbf{MLLM Selection.} GLMP requires an MLLM capable of processing a large set of representative patches from a WSI within a single prompt. Because this input typically includes \textasciitilde250 image patches, the MLLM must support a substantially larger visual-token capacity than is available in most current models. Many widely used general-purpose MLLMs \citep{openai2023gpt4v, anthropic2024claude3} and pathology-specific MLLMs such as SmartPath-R1 and CPath-Omni \citep{xu2025smartpath, sun2024cpathomni} offer limited vision context windows that cannot accommodate this input volume. Thus, we adopt Gemini 2.5 Pro \citep{comanici2025gemini} for all experiments, and evaluate Qwen3-VL-235B-A22B-Instruct \citep{qwen3technicalreport} and GLM-4.5V \citep{vteam2025glm45} in selected experiments to assess the sensitivity of GLMP to different MLLMs.

\section{Model}
\label{overview}

We introduce GLMP, a pathology model designed to generate generalizable patch-level histology embeddings that are robust to batch effects (\autoref{fig:schematic}). In contrast to standard vision-only PFMs, GLMP leverages an MLLM to derive text representations that serve as a middleman between the image patch and the numeric embedding. 

First, after partitioning a WSI into patches of size $128\times128\,\mu\mathrm{m}^2$ (approximately $256 \times 256$ pixels under $20\times$ magnification) using a non-overlapping sliding window, GLMP employs a PFM (Virchow2 \cite{zimmermann2024virchow2} in our implementation) to perform a within-sample clustering. This process identifies a set of distinct histology clusters and makes the analysis scalable by enabling the selection of a manageable subset of patches for MLLM prompting.

We then use an MLLM (Gemini 2.5 Pro \cite{comanici2025gemini} in our implementation) to generate text descriptions for the clusters of image patches. Guided by a single structured prompt per WSI (\ref{sec:standard_prompts}), the MLLM generates a distinct output for each of the histology clusters. This process directs the model to focus on histopathologic features that are relevant to the underlying biological characteristics, yielding semantic descriptions that minimize non-biological, TSI-specific artifacts (see \ref{sec:appendix_example_output} for examples). 

These cluster-level descriptions are processed by a text encoding model (Gemini Embedding \cite{gemini_embedding_2025} in our implementation) to produce a core embedding for each cluster. Patch-level PFM features are used to calculate soft-membership scores via a temperature-scaled softmax over cosine similarities between patch embeddings and cluster centroids. The final, unique embedding for each individual patch is then computed as a weighted sum of cluster embeddings. See \ref{sec:appendix_workflow} for the technical details of the full workflow.

\section{Experiments}

\subsection{Datasets and baselines}
\label{sec:setup}

We evaluated our approach on the following 5 datasets. Acknowledging the MLLM's unknown pretraining corpus, our evaluation uses datasets that, as originally published, lack the image-text pairs used for MLLM pretraining. To further ensure the data was truly unseen, we also included the AI4SkIN and TumSeg datasets, published after the MLLM (Gemini 2.5 Pro)'s knowledge cutoff date.
\underline{CAMELYON16}: Sentinel lymph node tissue samples (665586 patch-level images from 56 WSIs) with region-level annotations of metastatic tumor, acquired from 2 TSIs in the Netherlands (Radboud UMC, UMC Utrecht) \citep{camelyon16}. 
\underline{TCGA-LUSC}: Diagnostic slides (242740 patch-level images from 30 WSIs) of lung squamous cell carcinoma from The Cancer Genome Atlas \citep{tcga_lusc} with region-level annotations of invasive tumor \citep{loeffler_2021_5320076}, acquired from 3 TSIs: Mayo Clinic Rochester (MCR), the International Genomics Consortium (IGC), and Indivumed.
\underline{AI4SkIN}: Tissue samples (1426042 patch-level images from 60 WSIs) of cutaneous spindle cell neoplasms from 2 TSIs in Spain (HCUV and HUSC) \citep{ai4skin_2025}. 
\underline{TumSeg}: WSIs (127690 patch-level images from 38 WSIs) of cutaneous squamous cell carcinoma from the Histo-Miner project \citep{sancere2025histominer}, acquired from 2 TSIs in Cologne and Munich.
\underline{MSBCD}: Multi-Study Breast Cancer Dataset, a collection of five tissue samples (20110 patch-level images from 5 WSIs) curated from 3 independent breast cancer studies \citep{janesick2023high, andersson2021spatial, 10x_visium_breast_cancer}. 

Our model is compared with a panel of state-of-the-art PFMs, including Virchow2 \citep{zimmermann2024virchow2}, UNI2-h \citep{chen2024uni}, hibou-L \citep{nechaev2024hibou}, H-optimus-1 \citep{dippel2024rudolfv}, Phikon-v2 \citep{hua2024pathoduet}, Prov-GigaPath \citep{xu2024gigapath}, and CONCH (image encoder only) \citep{lu2024visual}.
We also include general-purpose vision models, such as DINOv2-base \citep{oquab2023dinov2} and ResNet-50 \citep{he2016resnet}, as well as general-purpose MLLMs (image encoders only), such as Qwen2.5-VL-7B-Instruct \citep{Qwen2VL} and Llama-3.2-11B-Vision \citep{grattonfiori2024llama}.  
See \ref{sec:appendix_baselines} for baseline model details.

\subsection{Multi-TSI embedding clustering}

\begin{figure*}[t]
\begin{center}
    \includegraphics[width=0.9\textwidth]{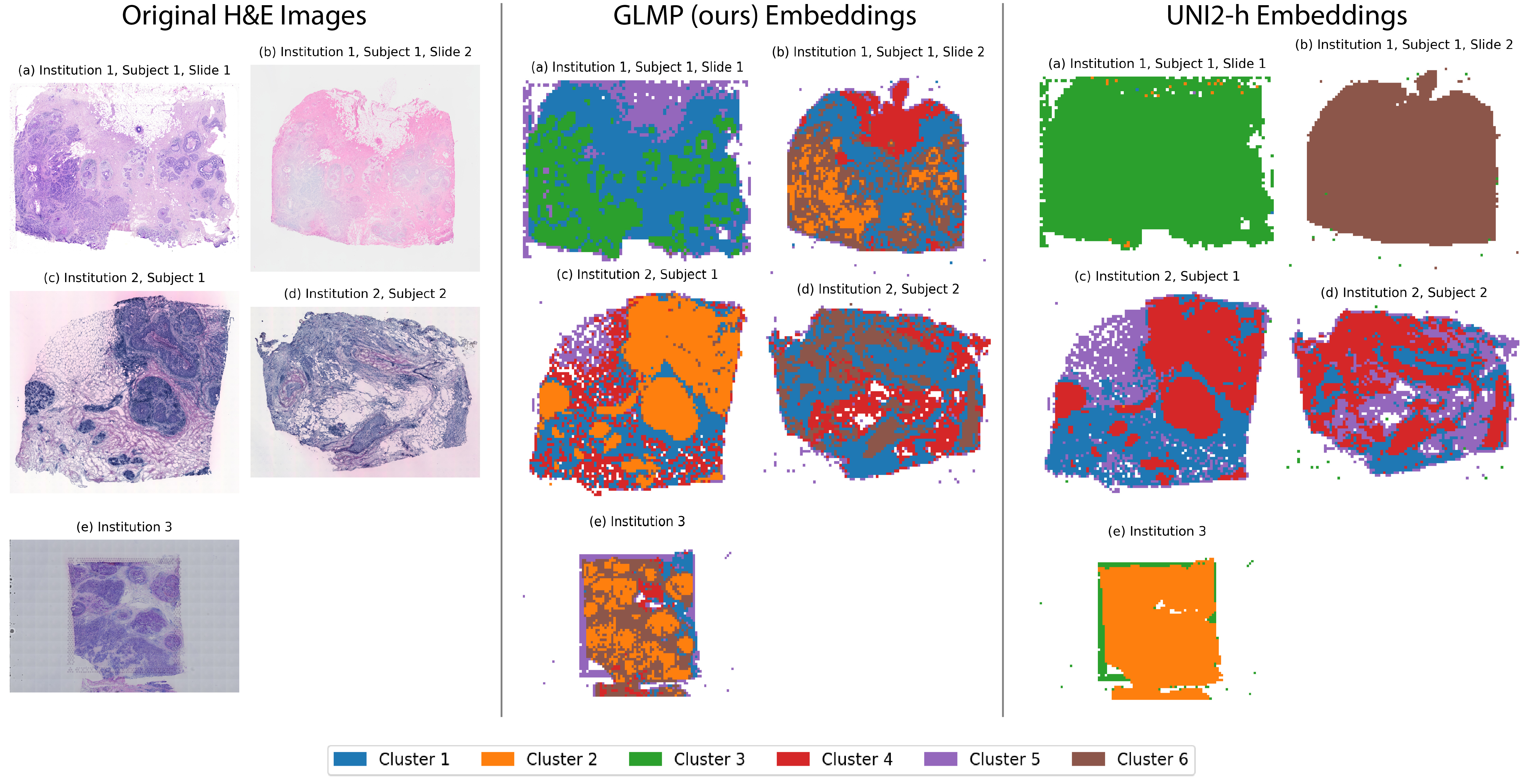}
\end{center}
  \caption{Cross-slide clustering results using $k$-means on patch embeddings from the Multi-Study Breast Cancer Dataset (MSBCD), which comprises 5 WSIs from 3 independent studies. Embeddings generated by different models are visualized alongside the original H\&E images to compare their alignment with biological tissue types. 
}
\label{fig:clustering}
\end{figure*}

To provide a visual overview of the robustness of GLMP compared with existing PFMs, we first demonstrate clustering results based on embeddings generated by GLMP and baseline PFMs on multi-TSI tissue slides in MSBCD. Clustering is performed at the patch level across 5 tissue slides using \(k\)-means. As shown in \autoref{fig:clustering}, GLMP clusters are consistently aligned with biological tissue types (e.g., brown for cancer, blue for stroma, red for adipose). The clusters are well-distributed across tissue slides and TSIs, indicating robustness against TSI-specific signatures. In contrast, embeddings by UNI2-h, one of the most downloaded PFMs on HuggingFace (122k as of November 2025), are predominantly driven by batch effects, as the three TSIs (with the exception of some noise in the image backgrounds) have three mutually exclusive sets of clusters  (e.g., green and brown for TSI 1; blue, red, purple for TSI 2; orange for TSI 3). Similar performance is observed in other baseline PFMs (\ref{sec:clustering_appendix}). These results show that the GLMP representation prioritizes genuine biological features over technical artifacts in tissues.

\subsection{Visualization of the embedding space}

\begin{figure*}[t]
\begin{center}
    \includegraphics[width=0.8\textwidth]{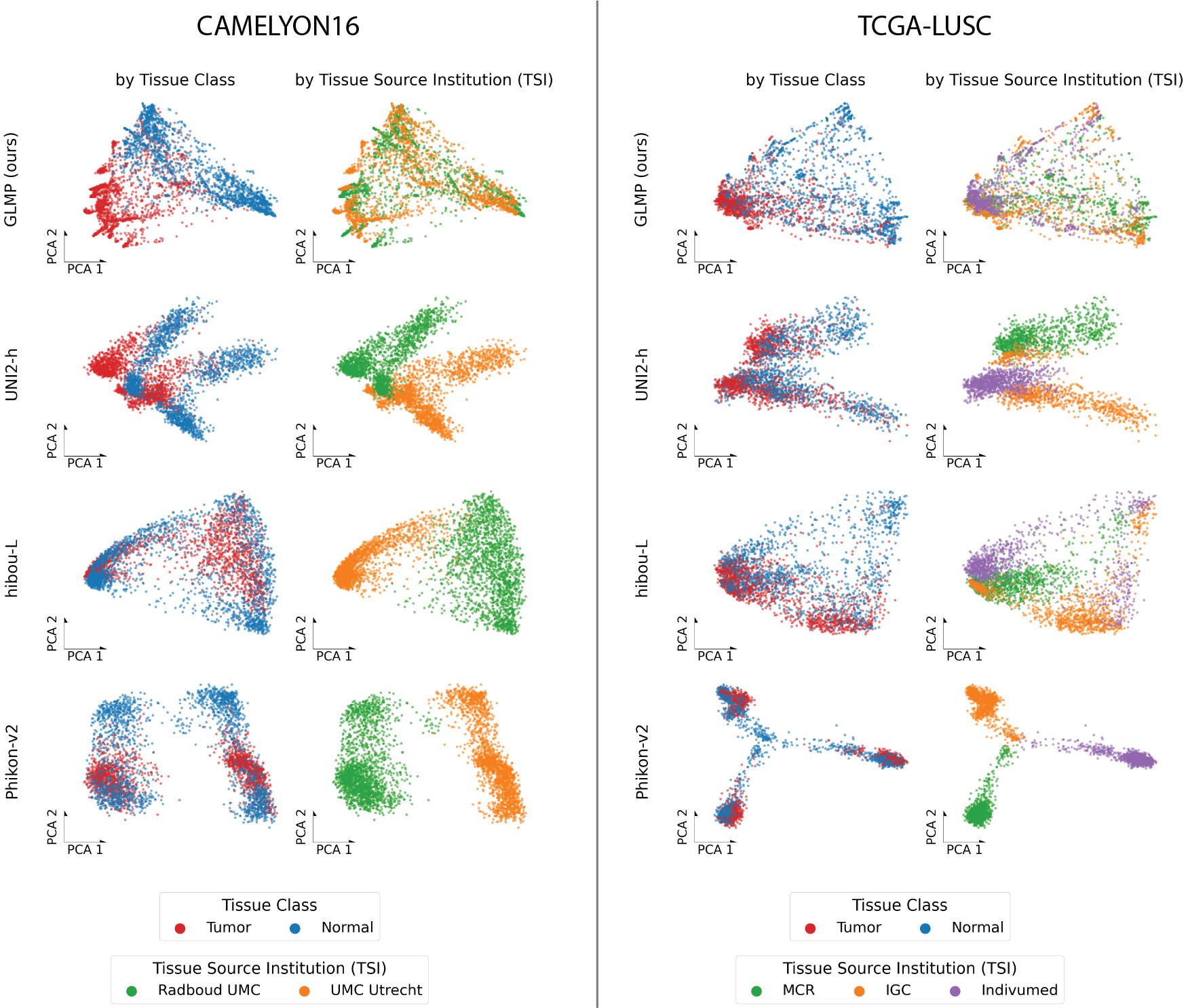}
\end{center}
\caption{PCA visualization of embeddings generated by GLMP and baseline PFMs on CAMELYON16 and TCGA-LUSC datasets, grouped by tissue class and TSI.}
\label{fig:dimension reduction}
\end{figure*}

To better understand the embedding space of different models, we perform principal component analysis (PCA) \citep{bishop2006pattern} and t-SNE \citep{maaten2008visualizing} to visualize the patch-level embeddings from CAMELYON16 and TCGA-LUSC. The resulting 2D projections are colored by both tissue class and TSIs to assess whether the latent structure reflects biological signals or batch effects. The dimension reduction results show that GLMP learns representations that differentiate distinct tissue types while remaining robust with respect to TSI-specific signatures (\autoref{fig:dimension reduction} \& \ref{sec:appendix_dim_reduction}). Across both datasets, GLMP embeddings show good separation between tumor and normal patches. When colored by TSI, the points are intermingled, suggesting minimal influence by TSI-specific signatures. By comparison, for the embeddings produced by baseline PFMs such as UNI2-h, hibou-L, and Phikon-v2, patches from different TSIs are clearly separated, and distances between TSIs tend to be larger than those between tissue classes. Together, these findings demonstrate that proximity in GLMP’s embedding space primarily reflects biological similarity rather than TSIs, while baseline PFMs are much more affected by batch effects.

\subsection{Cross-TSI generalization}
\label{main_sec:generalization}

An important clinical application of PFMs is the prediction of tissue classes. In practice, models are commonly trained on data from one or more TSIs and then deployed to new TSIs. Thus, we assess the cross-TSI generalization of GLMP compared with baseline PFMs by performing cross-TSI testing on the patch-level tissue classification task using CAMELYON16 and TCGA-LUSC. In each split of the cross-TSI testing, one TSI is selected to produce the testing image patches, while patches from the other TSIs are used for training. Embeddings are generated by the model for all the patches, and we then train a linear probe to classify tumor vs. normal. We also performed within-TSI prediction for the same task, where patches from all the TSIs are mixed together and randomly sampled to form the training-testing split, so as to assess how the model's performance is affected when testing is done on external instead of internal TSIs. Full experimental protocols are provided in \ref{sec:appendix_training}.

\begin{figure}[!ht]
\begin{center}
    \includegraphics[width=1\linewidth]{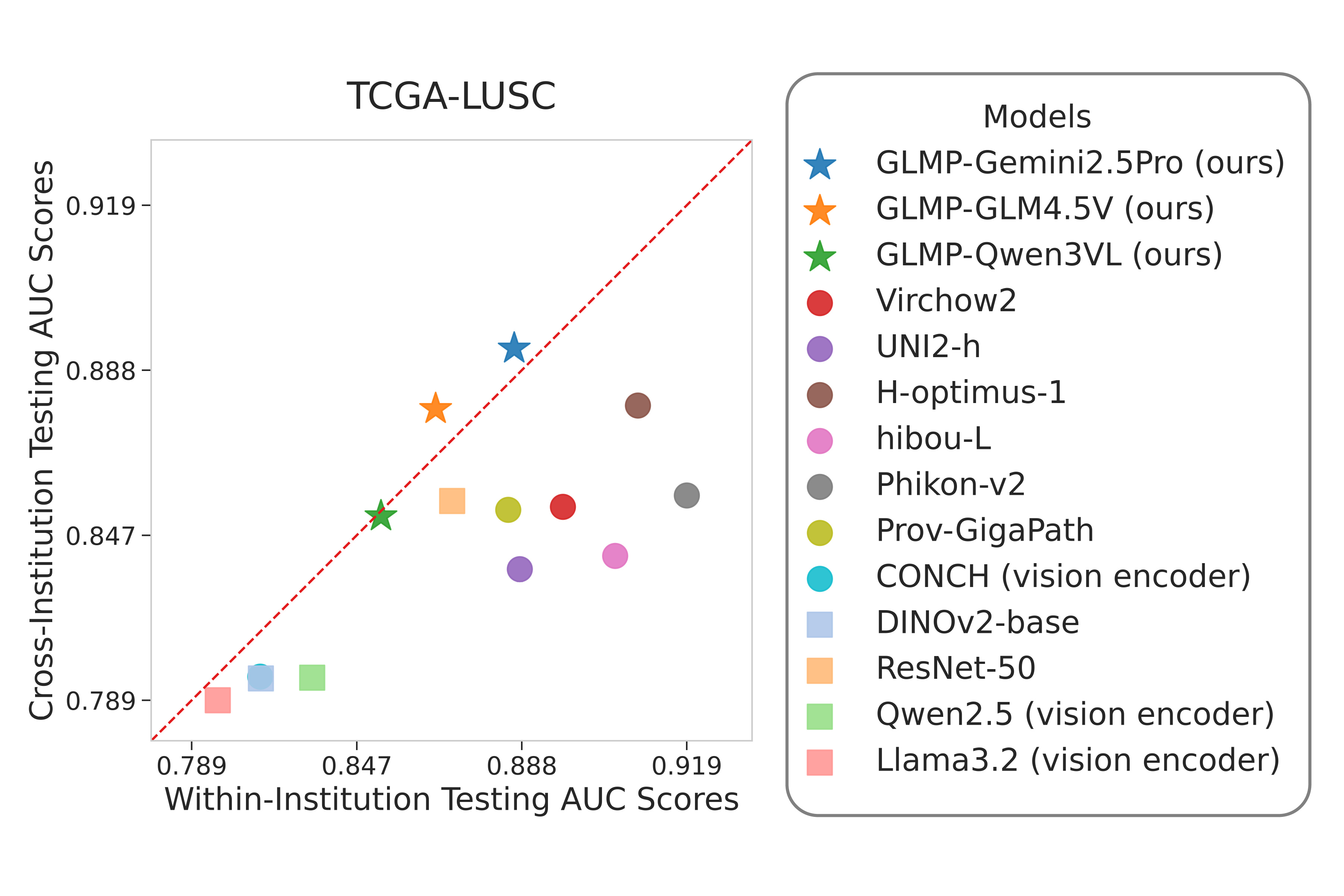}
\end{center}
  \caption{Accuracy of within-TSI testing (i.e. training data includes some WSIs from the testing TSIs) vs. cross-TSI testing (i.e. training does not include any WSIs from the testing TSIs) on TCGA-LUSC tissue classification task. Patches in training and testing sets are from disjoint WSIs, ensuring that patches from the same WSI do not appear in both sets. Points falling below the $y=x$ line indicate a performance drop in model generalization when new TSIs (not included during training) are used for testing. Accuracy is measured by AUC, with the axes transformed using -log(1-AUC) for clearer visualization scale. PFMs are marked with circles while general-purpose vision models are marked with squares.}
\label{generalization_vs_internal}
\end{figure}

The generalization results show that GLMP produces highly discriminative embeddings (\autoref{generalization_vs_internal} \& \autoref{generalization_vs_internal_cam16}). On TCGA-LUSC, while some baseline PFMs still achieve a higher accuracy than GLMP for within-TSI prediction, GLMP outperforms all baseline models in cross-TSI generalization, which highlights its robustness to TSI-specific artifacts. Compared with GLMP, general-purpose vision models (e.g. DINOv2-base, ResNet-50) and the vision encoders of general-purpose MLLMs (e.g. Qwen2.5-VL, Llama-3.2) show significantly lower performance in both within-TSI and cross-TSI testing, which indicates that these broad-domain models have limited applicability to histology images without further domain-specific adaptation. We also experimented with implementing GLMP using two open-weight MLLMs on this dataset, Qwen3-VL-235B-A22B-Instruct and GLM-4.5V, which yielded inferior performance compared to the original implementation of GLMP with Gemini 2.5 Pro, suggesting the gap between frontier proprietary MLLMs and currently available open-weight MLLMs.

To assess whether standard batch-effect mitigation strategies reduce the generalization gap, we augmented the baseline PFMs with Macenko normalization (image-level correction \citep{macenko2009method}) and Domain-Adversarial Neural Networks (DANN; feature-level alignment \citep{ganin2016dann}). Numeric results are reported in \autoref{cross_tsi_numeric}. We found that applying Macenko normalization yielded either marginal improvements or slight performance degradations on cross-TSI generalization for the baseline PFMs. Applying DANN-style feature alignment generally degraded performance across the PFMs. Overall, neither strategy substantially mitigated the cross-institutional performance drop.

Compared with the baseline PFMs, GLMP has the least reduction in performance between within-TSI and cross-TSI evaluation, which in practice means that GLMP can be deployed to new TSIs without the need for costly model retraining or fine-tuning, especially when no annotated data are available to fine-tune the model. This is highly relevant for institutions with limited case volumes or those contributing only occasional samples to multi-site studies, where the scarcity of labeled data makes local fine-tuning impractical.

\subsection{Robustness to TSI confounding}

A particularly dangerous case of batch effects is TSI confounding, where TSI-specific artifacts are correlated with true biological signals. Models might exploit these spurious correlations as predictive shortcuts, leading to biased predictions on new TSIs where the correlations do not hold. This kind of shortcut learning has been found to be a common pitfall for PFMs and natural vision models \cite{komen2024batch,hermann2023foundations}.
To evaluate the robustness of GLMP against TSI confounding, we follow a protocol adapted from \cite{komen2024batch} to create training sets with increasing correlations between the tissue class and TSI. We evaluate this on a CAMELYON16 subcohort and a TCGA-LUSC subcohort. For CAMELYON16, we construct three distinct training sets, ranging from a balanced split (50/50, no bias), to a moderately biased split (75/25), and to a fully biased split (100/0) where all normal tissue originated from one TSI (Radboud UMC) and all tumor tissue from the other (UMC Utrecht) (\autoref{tab:shortcut_data}). Similarly, for TCGA-LUSC, we introduce increasing levels of spurious correlations (50/50, 67/33, 83/17, and 100/0) across three TSIs (detailed in \ref{sec:tcga tsi confounding}). For both datasets, model performance is evaluated on a test set constructed with the direction of correlation reversed. This setup simulates a challenging real-world scenario where a model trained on biased data needs to generalize to an unbiased or reverse-biased setting. 

Since TSI-specific biases in WSIs are commonly addressed by stain normalization \cite{komen2024batch,lin2025impact,nguyen2025fmmap,yun2024exaonepath}, we provide baseline PFMs with Macenko \cite{macenko2009method} stain normalization. Additional experiments using Reinhard \cite{reinhard2001color} normalization yielded inferior results compared to Macenko and are therefore detailed in \ref{app:reinhard}.

\begin{table}[ht]
\footnotesize 
\centering
\caption{Dataset splits for the TSI confounding experiment on CAMELYON16. The training sets introduce spurious correlations between the tissue class and TSI, while the testing set is constructed with a reversed correlation.}
\label{tab:shortcut_data}
\begin{tabular}{@{}ll rrrr r@{}}
\toprule
& & \multicolumn{2}{c}{\textbf{Radboud UMC}} & \multicolumn{2}{c}{\textbf{UMC Utrecht}} & \\
\cmidrule(lr){3-4} \cmidrule(lr){5-6}
Split & Set & \makecell{\# \\Normal} & \makecell{\# \\ Tumor} & \makecell{\# \\ Normal} & \makecell{\# \\Tumor} & \makecell{Total \\Patches} \\
\midrule
\makecell{50/50 \\ (no bias)}  & Train & 1,600 & 1,600 & 1,600 & 1,600 & 6,400 \\
\makecell{75/25 \\ (low bias)}  & Train & 2,400 & 800   & 800   & 2,400 & 6,400 \\
\makecell{100/0 \\ (high bias)} & Train & 3,200 & 0     & 0     & 3,200 & 6,400 \\
\midrule
 & Test     & 0     & 800   & 800   & 0     & 1,600 \\
\bottomrule
\end{tabular}
\end{table}

\begin{figure*}[ht]
\begin{center}
    \includegraphics[width=0.7\textwidth]{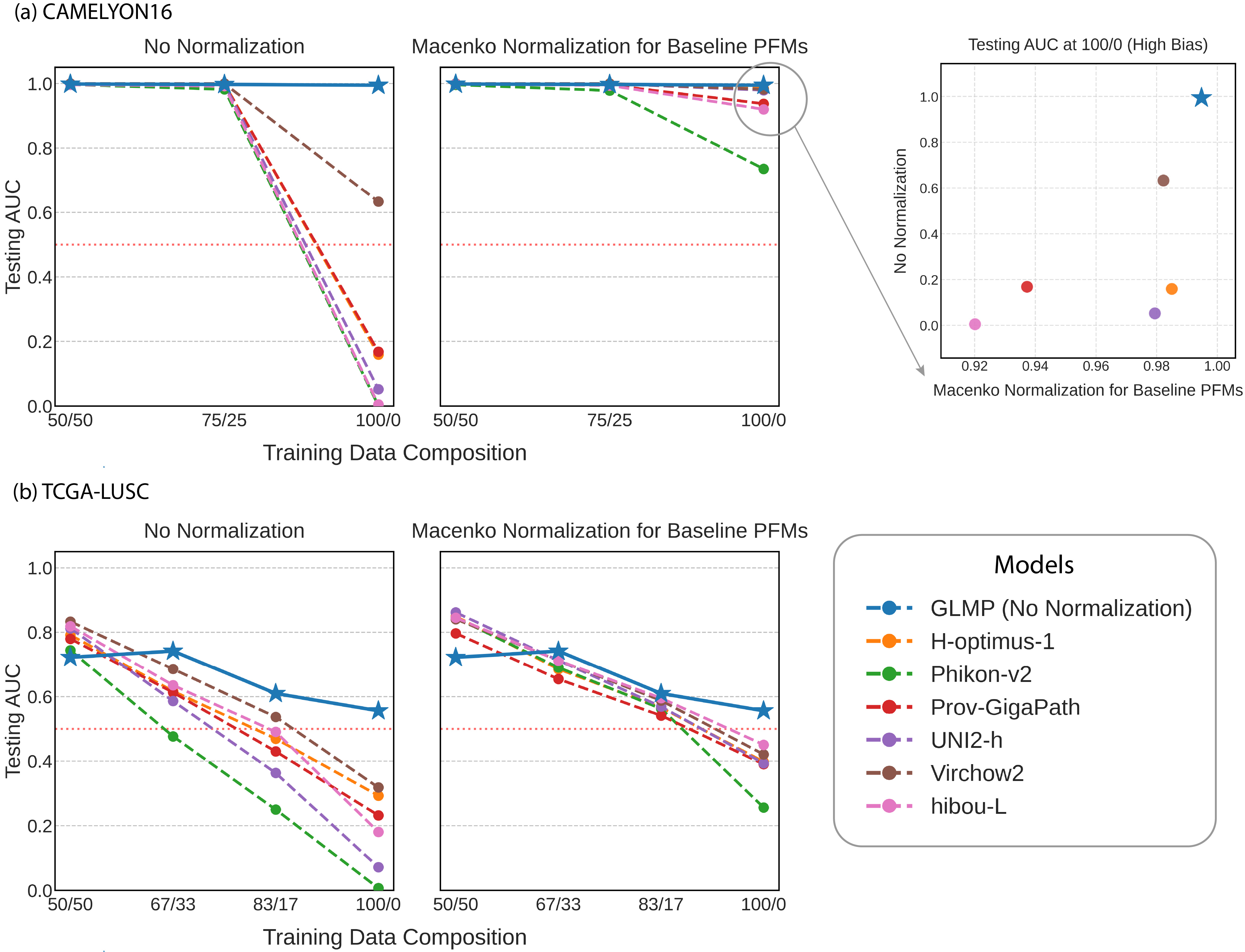}
\end{center}
  \caption{Tissue classification performance under increasing correlations between class label and TSI in (a) CAMELYON16 and (b) TCGA-LUSC. A single testing set with a reversed label-TSI correlation relative to the biased training sets is used for evaluation. Baseline PFMs are given the option of using Macenko stain normalization.}
\label{fig:shortcut}
\end{figure*}

\autoref{fig:shortcut} shows that GLMP is robust to TSI-based batch effects as a confounder. On CAMELYON16, GLMP maintains stable, near-perfect performance across all training conditions. Even in the most challenging scenario (100/0) where the training and testing data are fully biased in opposite directions, GLMP shows no reduction in prediction accuracy. On TCGA-LUSC, while all models exhibit performance degradation as confounding increases, GLMP experiences significantly less drop-off and remains the only model to achieve an AUC above 0.5 in the 100/0 case.

In contrast, without stain normalization, baseline PFMs perform well on the balanced splits but show severe degradation on the high-bias split. Their testing AUCs drop to near zero at the 100/0 split, suggesting that they rely on TSI signatures as a shortcut learning proxy for the tissue class. With Macenko normalization, the baseline PFMs show improved performance but they remain below GLMP. These findings indicate that while stain normalization helps mitigate batch effects across TSI for baseline PFMs, it does not fully resolve the issue. On the other hand, GLMP’s language-mediated representations are less susceptible to spurious correlations in the training data, enabling consistent generalization without the need for normalization. GLMP's generalizability under severely biased data conditions highlights its potential for reliable deployment in real-world clinical settings where data biases are common.

\subsection{Neighborhood-based Robustness Evaluation}
\label{sec:robustness_index}

To further quantify the degree to which representations capture genuine biological features over TSI-specific artifacts, we evaluate GLMP and the baseline PFMs on robustness-oriented downstream tasks using the robustness index (RI) \cite{komen2025towards}, which reflects the extent to which a patch's local neighborhood in the embedding space is driven by biological similarity rather than confounding TSI-specific artifacts. We conduct this evaluation on a balanced CAMELYON16 subcohort comprising normal and tumor samples from two TSIs (computational details in \ref{sec:appendix_training}). As shown in \autoref{tab:robustness_index}, GLMP achieves the highest RI among all the evaluated models, indicating that its embedding neighborhoods are primarily organized by biological tissue characteristics rather than TSI-specific artifacts. These results suggest that the local neighborhoods in their embedding spaces are largely influenced by TSI-specific signals rather than biological similarity. Overall, GLMP produces embeddings that more consistently reflect underlying tissue morphology, which is advantageous for downstream tasks that rely on biologically meaningful representation structure, such as image retrieval, unsupervised discovery of novel disease subtypes, and spatial omics prediction.

\subsection{TSI Prediction}
\label{sec:med_center_pred}
The ideal PFM should be invariant to non-biological technical variation and agnostic to TSIs. To quantify TSI-specific batch effects, we adopt the TSI prediction task as in \citet{komen2024batch}. Specifically, for each model, we train a linear probe to predict a patch's TSI from its embedding and evaluate its performance via cross-TSI testing (\ref{sec:appendix_training}). Lower TSI prediction accuracy indicates less potential influence by TSI-specific artifacts and thus better robustness to batch effects.

Most baseline PFMs achieve very high TSI prediction accuracy, often close to 100\%, indicating that their embeddings contain strong TSI-specific signatures.
In contrast, GLMP's TSI prediction accuracy is only slightly above random chance, suggesting that GLMP's embeddings are largely free of information about TSIs (\autoref{tab:med_center_pred}).
The results confirm that GLMP is more robust to batch effects and less likely to exploit TSI-specific artifacts as predictive shortcuts than existing PFMs.

\begin{table}[h!]
\begin{threeparttable}
\centering
\footnotesize 
  \caption{Model susceptibility to batch effects, measured by accuracy on the TSI prediction task using a linear probe. Lower accuracy indicates less potential influence by TSI-specific signatures and higher robustness to technical artifacts, with a truly TSI-agnostic model expected to perform no better than random chance. Standard deviations across cross-validation folds are reported in \ref{sec:TSI_pred_table}.}
\label{tab:med_center_pred}
\begin{tabular}{@{}lrrrr@{}}
\toprule
Model & CAM16 & TCGA-LUSC & AI4SkIN & TumSeg \\
\midrule
\makecell[l]{Completely random \\(ideal performance)} & 0.5000 & 0.3333 & 0.5000 & 0.5000 \\
  GLMP (ours) & \underline{0.6212} & \underline{0.4750} & \underline{0.5287} & \underline{0.6154} \\
Virchow2 & 0.9998 & 0.9600 & 0.9926 & 0.9209 \\
UNI2-h & 0.9999 & 0.9690 & 0.9965 & 0.9201 \\
hibou-L & 0.9995 & 0.9593 & 0.9983 & 0.8962 \\
H-optimus-1 & 0.9999 & 0.9794 & 0.9914 & 0.8896 \\
Phikon-v2 & 0.9999 & 0.9973 & 0.9894 & 0.9027 \\
Prov-GigaPath & 0.9997 & 0.9560 & 0.9893 & 0.9278 \\
CONCH & 0.9985 & 0.7222 & 0.9536 & 0.8559 \\
DINOv2-base & 0.9966 & 0.7335 & 0.9640 & 0.8395 \\
ResNet-50 & 0.9877 & 0.8316 & 0.9690 & 0.8030 \\
Qwen2.5-VL\tnote{a}& 0.9990 & 0.8124 & 0.9907 & 0.8825 \\
Llama-3.2\tnote{b} & 0.9976 & 0.6665 & 0.9813 & 0.8467 \\
\bottomrule
\end{tabular}
\begin{tablenotes}
    \item[a] Qwen2.5-VL-7B-Instruct
    \item[b] Llama-3.2-11B-Vision
\end{tablenotes}
\end{threeparttable}
\end{table}

We further assessed the preeminence and concentration of TSI-specific signatures among all the variations captured in the embeddings by examining the predictability of TSI by the top principal components (PCs) based on a protocol adapted from \cite{komen2024batch}. Since the top PCs reflect the directions of highest variance in the embeddings, this experiment evaluates how prominent the TSI-specific signatures are in the model feature space. As shown in \autoref{fig:pc_knn}, baseline PFMs achieve high TSI prediction accuracy with only a few PCs regardless of stain normalization, indicating that non-biological artifacts are among the dominant sources of variation in their embeddings. By contrast, GLMP's TSI prediction accuracy remains agnostic toward institutional differences even when using a high number of PCs. These results show that GLMP encodes substantially less TSI-specific signatures than existing PFMs.

\subsection{Ablation studies}

We conducted ablation studies to validate the contribution of key components in the GLMP framework, specifically the prompting strategy and the MLLM backbone.

First, we would like to understand why GLMP is robust to TSI-specific signatures. We hypothesize that the biology-focused prompt for the MLLM filters out technical artifacts in the text representations of the images. To validate this hypothesis, we modify the prompt to explicitly instruct the MLLM to comment on non-biological characteristics, such as staining intensity and color profile, in addition to biological features (\ref{sec:non_bio_prompts}). Then the resulting embeddings are applied to the TSI prediction task on CAMELYON16. When the text representations are polluted with descriptions about non-biological attributes, the resulting embeddings become much more predictive of TSI (\autoref{tab:prompt_ablation}). This result indicates that by focusing the MLLM on biological features, GLMP effectively filters out technical artifacts, making the embeddings more robust to batch effects across TSIs. The language mediation layer in GLMP serves as a gatekeeper that determines what information is allowed to pass from the image to the embeddings based on the instructions in the MLLM prompt.

To further assess the framework's stability, we evaluated its sensitivity to component choices. We tested alternative clustering configurations (varying both the algorithm and the number of clusters $k$) and different text encoders, including domain-specific models like PubMedBERT. GLMP's performance remained highly consistent across all variations, demonstrating that its robustness is intrinsic to the language-mediated design rather than specific component tuning (full results in \ref{tab:sensitivity_ablation}).

Finally, we investigate how the choice of MLLM affects GLMP's performance. We replace Gemini 2.5 Pro, the default MLLM in GLMP, with Gemini 2.0 Flash, a faster model with lower reasoning strength and overall intelligence. Both implementations are applied to the CAMELYON16 tissue classification task. As shown in \autoref{tab:mllm_benchmark}, using Gemini 2.0 Flash leads to a drop in classification accuracy compared with Gemini 2.5 Pro. This finding indicates that the MLLM's overall capability does impact GLMP's performance. Furthermore, we evaluate the necessity of the full GLMP pipeline against direct "MLLM-only" classification. We test baselines where the MLLM is prompted to directly predict tumor presence using either single patches or the same representative patch clusters as GLMP (see \ref{sec:direct_prompts_patch} \& ~\ref{sec:direct_prompts_cluster} for details). While direct cluster-level classification achieved a respectable accuracy of 0.86 (surpassing the single-patch baseline of 0.70), GLMP yields the highest performance (0.91). This confirms that generating intermediate semantic embeddings captures more robust diagnostic signals than relying on the MLLM's immediate zero-shot binary decisions.

\begin{table}[ht]
\small
\centering
  \caption{Tissue classification accuracy of different MLLM-based methods.}
\label{tab:mllm_benchmark}
\begin{tabular}{@{}lr@{}}
\toprule
Method & \makecell{Tissue classification\\ accuracy} \\
\midrule
  GLMP (Gemini 2.5 Pro) & 0.91 \\
  GLMP (Gemini 2.0 Flash) & 0.74 \\
  MLLM-only (Gemini 2.5 Pro, by cluster) & 0.86 \\
  MLLM-only (Gemini 2.5 Pro, by patch) & 0.70 \\
  
\bottomrule
\end{tabular}
\end{table}

\section{Conclusion}

In this work, we have introduced GLMP, a pathology model with robustness to batch effects and generalizability across TSIs. A core innovation of GLMP is the use of an MLLM to translate histology images into biology-focused text intermediaries before encoding them numerically. A key finding in our experiments is that when used properly, vision-language models pretrained on massive general-purpose data can effectively extract pathologically relevant features while filtering out TSI-related artifacts. Future work will extend GLMP beyond H\&E-stained WSIs to other staining and imaging modalities, tailor it to disease-specific applications, and develop theoretical analyses of its generalization across TSIs.

\section*{Acknowledgements}
D.Z. is supported by P30ES010126 and hardware (A100 80GB $\times$ 8) provided by an NVIDIA Academic Grant. Y.L. is supported by R01AG079291 and R01AG085581.

\onecolumn
\appendix
\renewcommand{\thesection}{Suppl.~\Alph{section}}
\renewcommand{\thefigure}{S\arabic{figure}}
\renewcommand{\thetable}{S\arabic{table}}
\renewcommand{\theequation}{S\arabic{equation}}

\section{GLMP Implementation Details}
\label{sec:appendix_workflow}

\paragraph{Foreground tissue detection.} Each whole-slide image (WSI) is tiled into non-overlapping square patches of a fixed physical size (e.g., \(128\times128\,\mu\mathrm{m}\)). We derive an adaptive tissue mask by converting a downsampled WSI thumbnail to the Hue, Saturation, and Value (HSV) color space and applying Otsu's method to the non-zero values of each channel~\citep{schreiber2024rapid}. Patches are retained for subsequent analysis only if their masked tissue area exceeds a minimum threshold of 1\%.

\paragraph{Patch clustering for efficient MLLM prompting.} 

We use a cluster-based strategy to ensure efficiency of the MLLM API calls. For each WSI, foreground patches are first clustered into $k$ histology patch groups using \(k\)-means based on their similarity in the feature space of a vision encoder backbone. Since within-slide clustering analysis by definition is invariant to slide-level (and thus TSI-level) batch effects, which are constant for any single tissue slide, this clustering step is commonly used for efficiently organizing within-slide patches into histopathologically representative groups \cite{ding2024multimodal,wang2023retccl,shaban2025foundation}. Existing works find this single-slide clustering process to be stable with respect to model choice and the number of clusters when $k$ is between 10 and 25 per WSI \citep{claudio2024mapping,alfasly2025validation,zhang2024inferring,shaban2025foundation,ge2025deep}. We thus set the number of clusters to $k=10$ in our experiments to control MLLM context window usage, since long contexts can lead to performance degradation \citep{liu2023lost,an2024make,jha2024characterizing}. The clustering serves only as a preprocessing step for efficient API usage and does not contribute to the final embeddings, which are generated entirely through the MLLM-text encoder pipeline. After clustering, a soft probabilistic membership score is then computed for each patch to every cluster via a temperature-scaled softmax over their cosine similarities:

\begin{equation}
  \label{eq:soft_assignment}
p_{ij} = \frac{\exp\!\left(\langle \hat{\mathbf{x}}_{i}, \hat{\boldsymbol{\mu}}_{j}\rangle/\tau\right)}
{\sum_{\ell=1}^{k} \exp\!\left(\langle \hat{\mathbf{x}}_{i}, \hat{\boldsymbol{\mu}}_{\ell}\rangle/\tau\right)}
\end{equation}
where $\langle \cdot , \cdot \rangle$ denotes cosine similarity and $\tau$ is the softmax temperature, which is set to $\tau=0.1$ as recommended in temperature-tuning literature \citep{agarwala2020temperature,ferrante2023vaesim}. These probabilities are used to select $m=25$ representative patches per cluster for the MLLM API call.

\paragraph{Semantic description generation.}
Biologically focused semantic descriptions are generated for all histology patch groups in an API call to the MLLM (default: Gemini 2.5 Pro \citep{comanici2025gemini}). The request consists of a structured prompt (\ref{sec:standard_prompts}) followed by representative patches from each group, with images placed immediately after their textual identifiers. To induce a deterministic and structured output, the generation temperature is set to 0.0, and the model is instructed to return a single JSON object covering all groups. The process is repeated 10 times to form an ensemble of outputs, allowing minor textual variations from the inherent stochasticity of the MLLM to be averaged out in the subsequent embedding step. 

\paragraph{Patch-level embedding generation.} The structured MLLM descriptions for the $k$ histology patch groups are encoded into $k$ corresponding semantic embeddings using a text embedding model (Gemini Embedding \citep{gemini_embedding_2025}). To reduce randomness in outputs due to the inherent stochastic properties of MLLMs, the embeddings from 10 repeated MLLM runs for each group are then averaged to yield a single embedding $\mathbf{s}_j$ for each group $j$. The final GLMP embedding for an individual patch, $\mathbf{z}_i$, is then computed as a weighted sum of these $k$ group embeddings using the patch's soft probabilistic membership score $p_{ij}$ from (\autoref{eq:soft_assignment}) as the weight:
\begin{equation}
\mathbf{z}_i = \sum_{j=1}^{k} p_{ij} \mathbf{s}_j
\end{equation}

\clearpage
\section{Experimental details}
\label{sec:appendix_training}

\paragraph{Data preprocessing.} For all datasets, image patches are extracted to represent a physical area of \(128\times128\,\mu\mathrm{m}\) at an effective 20x magnification (\(\approx0.5\,\mu\mathrm{m}/pixel\)). The exception is the TumSeg dataset, which is available only at a low resolution (1.25x), from which we extract 256µm x 256µm patches. For CAMELYON16 and TCGA-LUSC, we select foreground patches located entirely within an annotated region ("Tumor") or entirely outside of any annotated region ("Normal"). 

\paragraph{Linear probe and training.}
We train a single linear layer on frozen embeddings, using the Adam optimizer with a learning rate of \(10^{-4}\) and cross-entropy loss. Training is performed with a batch size of 256 patches for 20 epochs, and we use the weights from the final epoch for evaluation.

\paragraph{Evaluation protocols.}
\begin{itemize}[leftmargin=*]
\item \textbf{Cross-TSI testing:} We evaluate generalization across TSIs by training and testing on slides from different TSIs. For this task, patch-level datasets are constructed by randomly sampling up to 2,000 tumor or normal patches from each slide. On the CAMELYON16 dataset, which includes 2 TSIs, models are trained on one TSI and tested on the other, and vice-versa. For the TCGA-LUSC dataset, we employ a leave-one-TSI-out testing, training on two TSIs and testing on the third, with the held-out TSI in rotation.

\item \textbf{Within-TSI testing:} We pool slides from all TSIs and use 5-fold splitting for creating the training and testing sets. Splits are created at the slide level so that no slide appears in more than one fold, and stratification maintains a balanced distribution of TSIs across folds. The patch sampling procedure (up to 2,000 patches per slide) is identical to that used in the cross-TSI experiments.

\item \textbf{TSI Confounding:} We adopt the protocol from \citet{komen2024batch} using a 56-slide subcohort from CAMELYON16. From each slide, 200 patches are sampled, with tumor patches drawn from metastatic slides and normal patches only from non-metastatic slides. This design yields three training splits with increasing levels of label–site correlation (50/50 no bias, 75/25 low bias, and 100/0 high bias), as summarized in \autoref{tab:shortcut_data}. A linear probe is then trained on each biased split and evaluated on a test set constructed with the opposite correlation.

\item \textbf{Neighborhood-based Robustness Evaluation:} Following \cite{komen2025towards}, we evaluated the Robustness Index (RI) on a strictly balanced $2\times2$ subcohort of CAMELYON16. We selected 10 slides for each combination of tissue class (Normal vs. Tumor) and TSI (Radboud UMC vs. UMC Utrecht), totaling 40 slides, and sampled 200 patches per slide. We computed the $k$-nearest neighbors using cosine similarity on $L_2$-normalized embeddings, dynamically masking patches from the same WSI as the query patch to ensure cross-slide retrieval. The neighborhood size $k$ was selected by first determining the optimal $k$ for each model via leave-one-slide-out cross-validation accuracy on the biological class, and then adopting the global median $k$ across all models for the final RI calculation.

\item \textbf{TSI Prediction:} We quantify site-specific artifacts by training a linear probe to predict the acquisition site across four datasets: CAMELYON16, TCGA-LUSC, AI4SkIN, and TumSeg. For each dataset, we sample up to 2,000 patches per slide and perform a 5-fold combined-TSI cross-testing. We report the mean prediction accuracy and standard deviation across folds.

\item \textbf{k-NN on Principal Components:} Following the protocol of \citet{komen2024batch}, we evaluate site-specific signal in the TCGA-LUSC and AI4SkIN embeddings. Using 5-fold cross-testing at the slide level, we train a \(k\)-nearest neighbors classifier (\(k=5\), cosine similarity) on the top \(l\) principal components, with \(l\) varying from 1 to 50. This assesses how much site information is captured in the dimensions of highest variance.

\end{itemize}

\clearpage
\section{Cross-TSI generalization}
\label{sec:appendix_cross_tsi}

\begin{figure}[ht]
\begin{center}
    \includegraphics[width=0.5\linewidth]{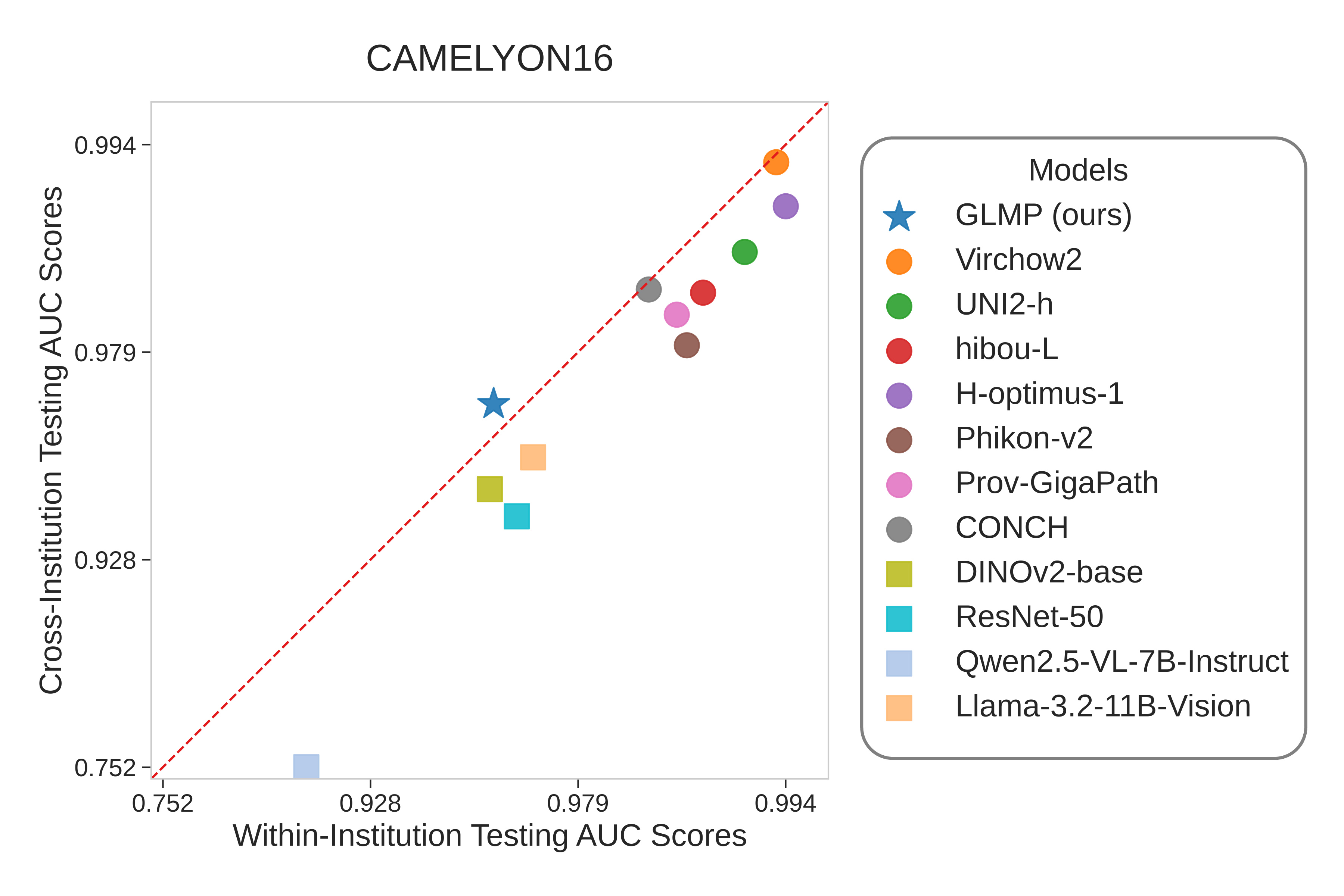}
  \caption{Accuracy of within-TSI testing (i.e. training data includes some WSIs from the testing TSIs) vs. cross-TSI testing (i.e. training does not include any WSIs from the testing TSIs) on CAMELYON16 tissue classification task. Patches in training and testing sets are from disjoint WSIs, ensuring that patches from the same WSI do not appear in both sets. Points falling below the $y=x$ line indicate a performance drop in model generalization when new TSIs (not included during training) are used for testing. Accuracy is measured by AUC, with the axes transformed using -log(1-AUC) for clearer visualization scale. PFMs are marked with circles while general-purpose vision models are marked with squares.
  }
  \label{generalization_vs_internal_cam16}
\end{center}
\end{figure}

\begin{table}[h]
    \centering
    \caption{AUC results for tissue classification under within-TSI and cross-TSI evaluation on TCGA-LUSC and CAMELYON16, corresponding to the correlation plots in \autoref{generalization_vs_internal} and \autoref{generalization_vs_internal_cam16}, including baseline PFMs and variants with Macenko normalization and DANN alignment.}
    \label{cross_tsi_numeric}
    \begin{tabular}{@{}lrrrr@{}}
    \toprule
    & \multicolumn{2}{c}{CAMELYON16} & \multicolumn{2}{c}{TCGA-LUSC} \\
    \cmidrule(lr){2-3} \cmidrule(l){4-5}
    Model & Within-TSI AUC & Cross-TSI AUC & Within-TSI AUC & Cross-TSI AUC \\
    \midrule
    GLMP (ours)            & 0.9653 & 0.9714 & 0.8819 & 0.8913 \\
    Virchow2               & 0.9935 & 0.9932 & 0.8968 & 0.8548 \\
    UNI2-h                 & 0.9922 & 0.9885 & 0.8878 & 0.8364 \\
    hibou-L                & 0.9900 & 0.9853 & 0.9066 & 0.8405 \\
    H-optimus-1            & 0.9939 & 0.9912 & 0.9106 & 0.8805 \\
    Phikon-v2              & 0.9890 & 0.9799 & 0.9187 & 0.8580 \\
    Prov-GigaPath          & 0.9883 & 0.9832 & 0.8853 & 0.8540 \\
    CONCH                  & 0.9862 & 0.9856 & 0.8154 & 0.7988 \\
    DINOv2-base            & 0.9645 & 0.9526 & 0.8157 & 0.7981 \\
    ResNet-50              & 0.9698 & 0.9443 & 0.8724 & 0.8564 \\
    Qwen2.5-VL-7B-Instruct & 0.8943 & 0.7517 & 0.8330 & 0.7984 \\
    Llama-3.2-11B-Vision   & 0.9726 & 0.9607 & 0.7998 & 0.7895 \\
    \midrule
    \multicolumn{5}{l}{\textbf{Macenko Normalization}} \\
    \midrule
    Virchow2               & 0.9872 & 0.9886 & 0.8928 & 0.8699 \\
    UNI2-h                 & 0.9916 & 0.9894 & 0.8899 & 0.8685 \\
    hibou-L                & 0.9873 & 0.9859 & 0.8958 & 0.8600 \\
    H-optimus-1            & 0.9927 & 0.9910 & 0.8978 & 0.8767 \\
    Phikon-v2              & 0.9862 & 0.9811 & 0.9020 & 0.8477 \\
    Prov-GigaPath          & 0.9845 & 0.9815 & 0.8843 & 0.8650 \\
    \midrule
    \multicolumn{5}{l}{\textbf{DANN}} \\
    \midrule
    Virchow2               & 0.9921 & 0.9861 & 0.8858 & 0.8442 \\
    UNI2-h                 & 0.9912 & 0.9821 & 0.8834 & 0.8296 \\
    hibou-L                & 0.9892 & 0.9869 & 0.9088 & 0.8554 \\
    H-optimus-1            & 0.9930 & 0.9901 & 0.8935 & 0.8550 \\
    Phikon-v2              & 0.9868 & 0.9759 & 0.9130 & 0.8500 \\
    Prov-GigaPath          & 0.9851 & 0.9737 & 0.8744 & 0.8549 \\
    \bottomrule
    \end{tabular}
\end{table}

\clearpage
\section{Baseline model configurations}
\label{sec:appendix_baselines}
Our comparative analysis involved 12 foundation models, encompassing pathology-specific vision encoders, general-purpose vision models, and MLLMs (Table \ref{tab:foundation_models}). All pre-trained models were obtained from the Hugging Face Hub, except for the ResNet-50 baseline from torchvision. For each model, the feature extraction procedure was tailored to its specific architecture and the authors’ recommendations. For Virchow2, we concatenated the classification token with the mean-pooled representation of all patch tokens. For hibou-L, we utilized the model's \texttt{pooler\_output}, while for Phikon-v2 and the DINOv2 vision encoder, we extracted the final hidden state of the classification token. For UNI2-h, H-optimus-1, and the Prov-GigaPath tile encoder, we used the single feature vector returned directly by a forward pass. For the contrastive vision-language model CONCH, we obtained features from its vision encoder prior to the final contrastive projection layer. The ResNet-50 baseline was represented by its final global average pooling layer. For the MLLMs, we investigated two types of semantic embeddings: features from the vision projection layer, and contextualized features from the full model's last hidden layer. Accordingly, for Qwen2.5-VL-7B-Instruct, we obtained the embeddings by mean-pooling the output of its \texttt{get\_image\_features} function. For Llama-3.2-11B-Vision, we obtained the contextualized embeddings by invoking the model's standard forward pass with a minimal, non-informative text prompt and then mean-pooling the representations from the last hidden layer.

\begin{table*}[ht]
\centering
\caption{Overview of the foundation models benchmarked in this study.}
\label{tab:foundation_models}
\resizebox{\textwidth}{!}{%
\begin{tabular}{@{}lllll@{}}
\toprule
\textbf{Model Repository ID} & \textbf{Architecture (Param.)} & \textbf{Pre-training Data} & \textbf{Dim.} & \textbf{Reference} \\
\midrule
\multicolumn{5}{c}{\textbf{Pathology-Specific Vision Encoders}} \\
\midrule
paige-ai/Virchow2 & ViT-H/14 (632M) & 3.1M WSIs & 2560 & \cite{zimmermann2024virchow2} \\
MahmoodLab/UNI2-h & ViT-H/14 (681M) & 350k WSIs & 1536 & \cite{chen2024uni} \\
histai/hibou-L & ViT-L/14 (304M) & 1.14M WSIs & 1024 & \cite{nechaev2024hibou} \\
bioptimus/H-optimus-1 & ViT-g/14 (1.1B) & 1M WSIs & 1536 & \cite{hoptimus1} \\
owkin/phikon-v2 & ViT-L/16 (304M) &  58.4K WSIs & 1024 & \cite{filiot2024phikon} \\
prov-gigapath/prov-gigapath & ViT-g/14 (1.1B) & 171k WSIs & 1536 & \cite{xu2024gigapath} \\
MahmoodLab/CONCH & ViT-B/16 (86M) & 1.17M image-text pairs & 512 & \cite{lu2024visual} \\
\midrule
\multicolumn{5}{c}{\textbf{General-Purpose Vision Encoders}} \\
\midrule
facebook/dinov2-base & ViT-B/14 (86M) & LVD-142M & 768 & \cite{oquab2023dinov2} \\
torchvision/resnet50 & ResNet-50 (25.6M) & ImageNet-1K & 2048 & \cite{he2016resnet} \\
\midrule
\multicolumn{5}{c}{\textbf{Multi-Modal Language Models (MLLMs)}} \\
\midrule
Qwen/Qwen2.5-VL-7B-Instruct & MLLM (7B) & Image-text/documents/video & 3584 & \cite{Qwen2VL} \\ 
meta-llama/Llama-3.2-11B-Vision & MLLM (11B) & 6B image-text pairs & 4096 & \cite{grattonfiori2024llama} \\ 
\bottomrule
\end{tabular}%
}
\end{table*}

\clearpage
\section{Additional experimental results}

\subsection{Multi-TSI Embedding Clustering}
\label{sec:clustering_appendix}

\begin{figure}[ht]
\begin{center}
    \includegraphics[width=1\linewidth]{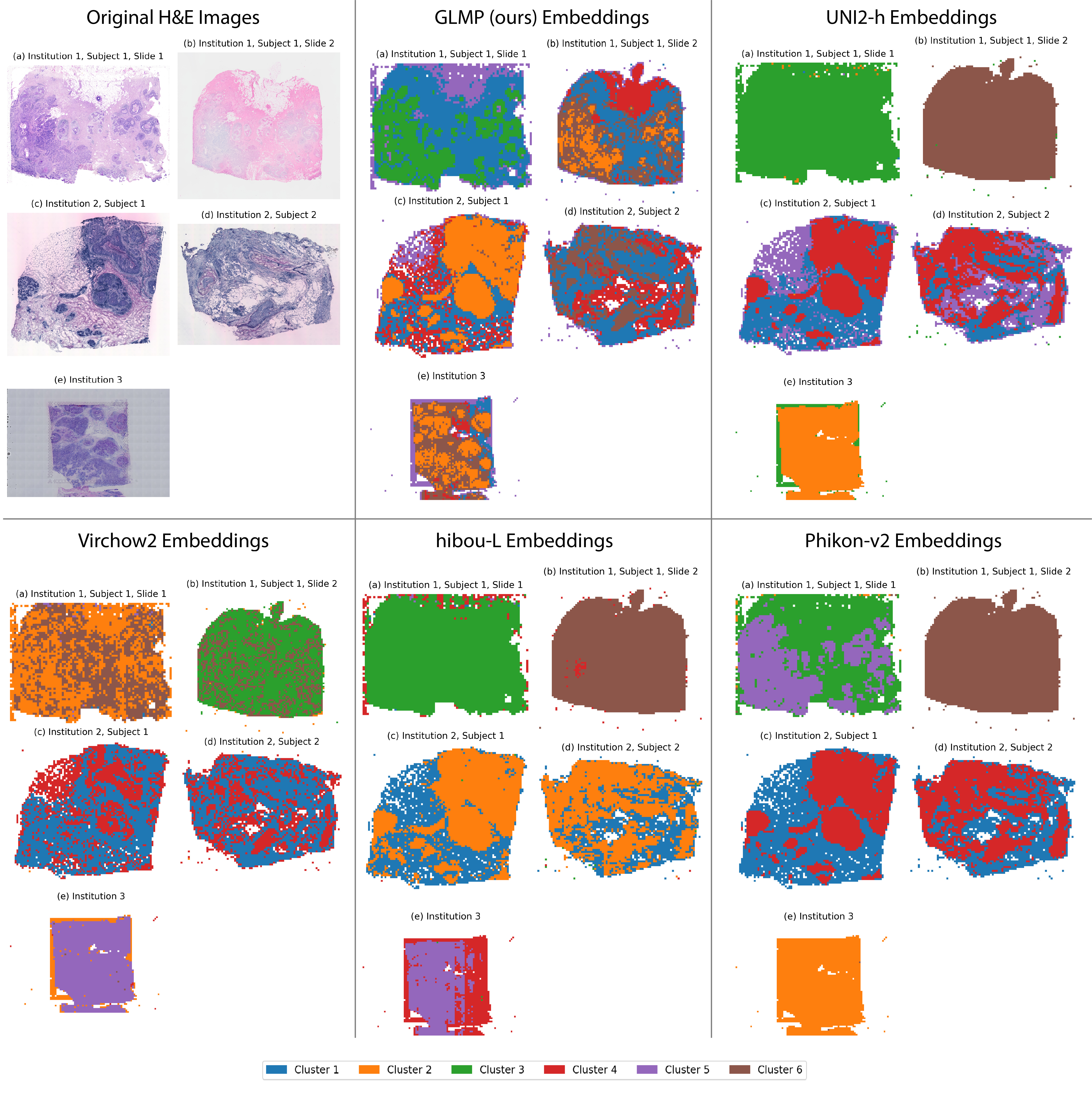}
  \caption{Clustering results on the MSBCD dataset using \(k\)-means on histology image patch embeddings generated by different models.
  }
\end{center}
\end{figure}

\clearpage
\subsection{Visualization of embedding spaces}
\label{sec:appendix_dim_reduction}

\begin{figure}[ht]
\begin{center}
    \includegraphics[width=0.9\linewidth]{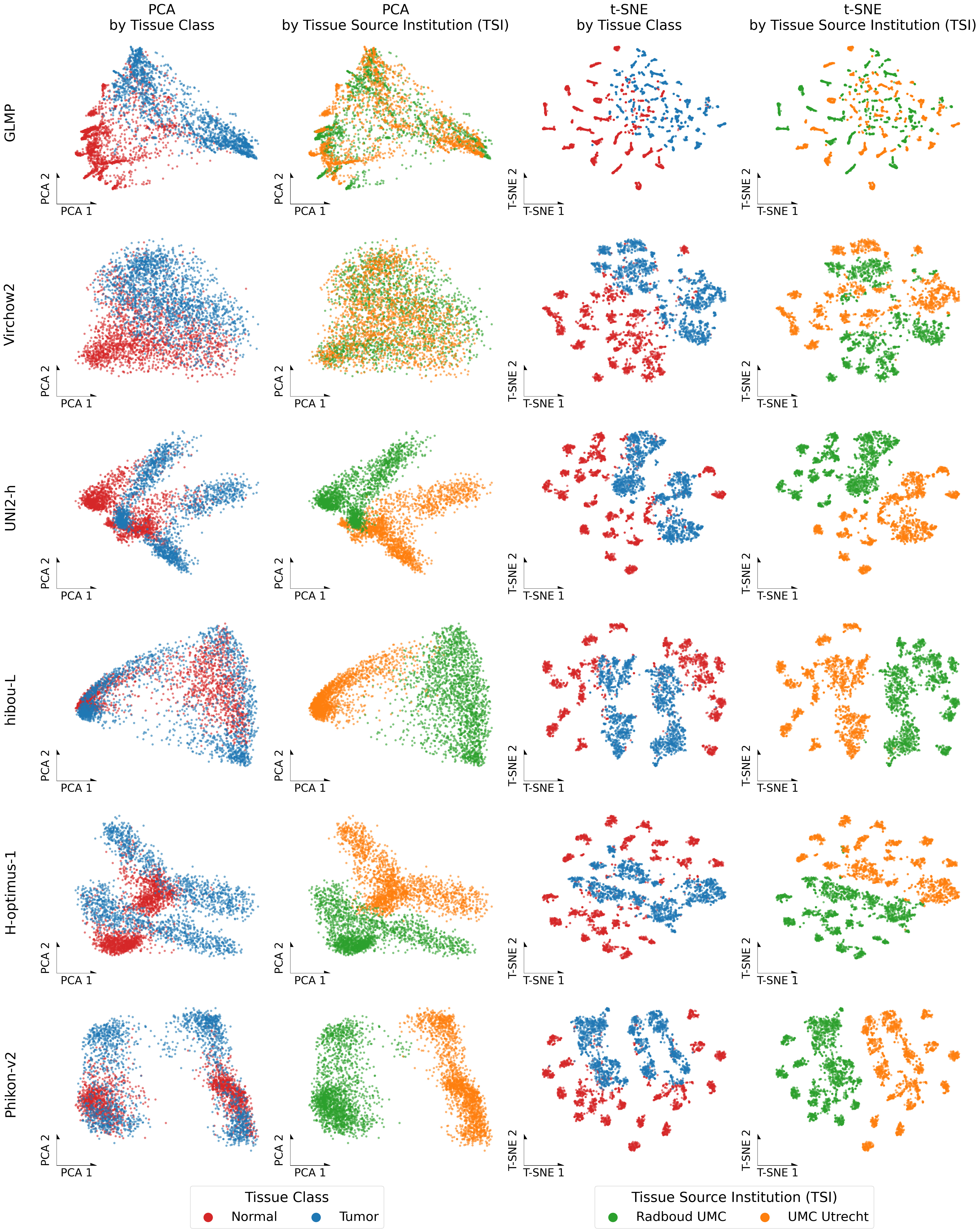}
\caption{Dimension reduction of pathology model embeddings on the CAMELYON16 Dataset.}
\end{center}
\end{figure}

\clearpage
\begin{figure}[t!]
\ContinuedFloat
\begin{center}
    
    \includegraphics[width=0.9\linewidth]{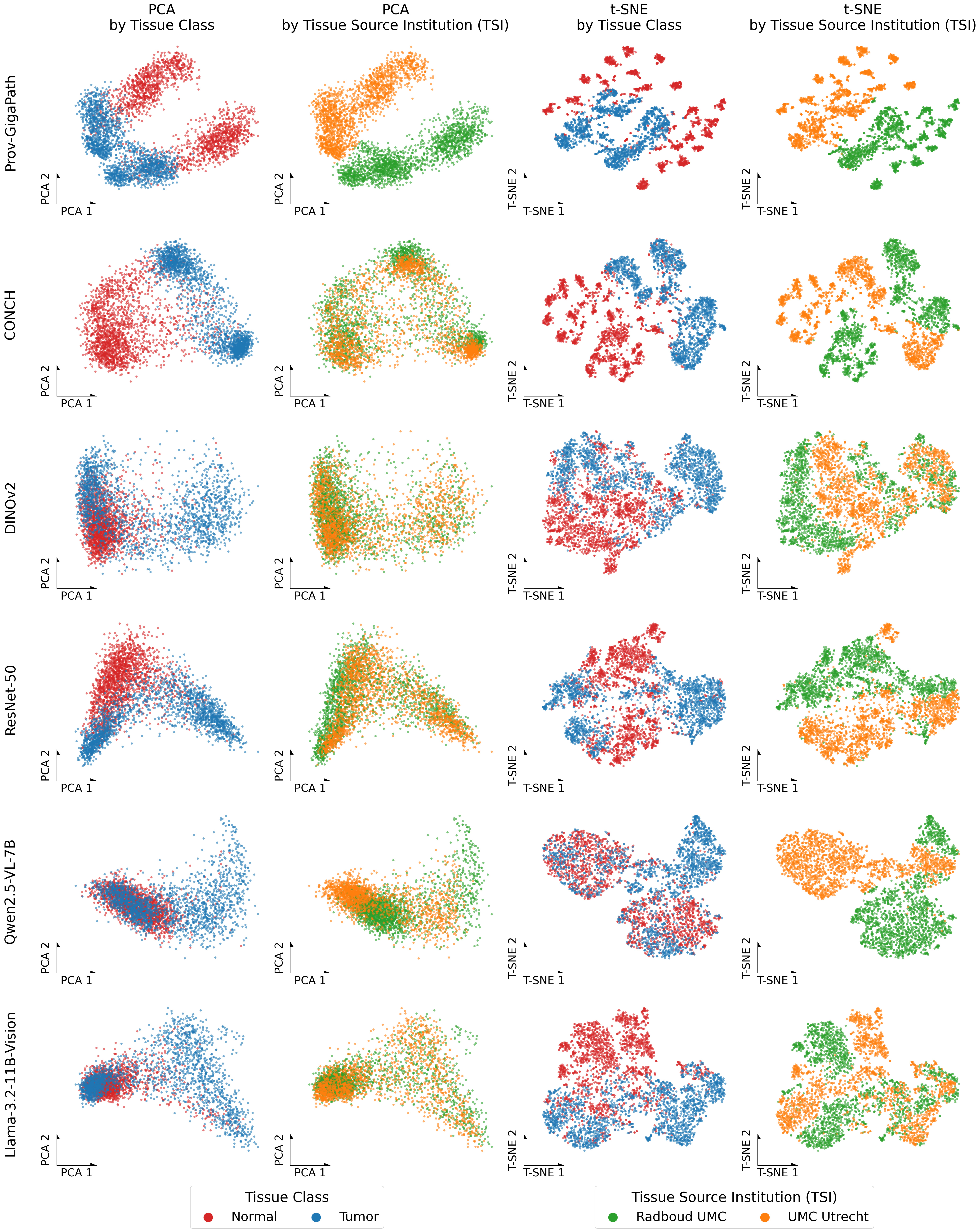}
\end{center}
\caption{(Continued) Dimension reduction of pathology model embeddings on the CAMELYON16 Dataset.}
\end{figure}

\clearpage
\begin{figure}[t!]
\begin{center}

    \includegraphics[width=0.9\linewidth]{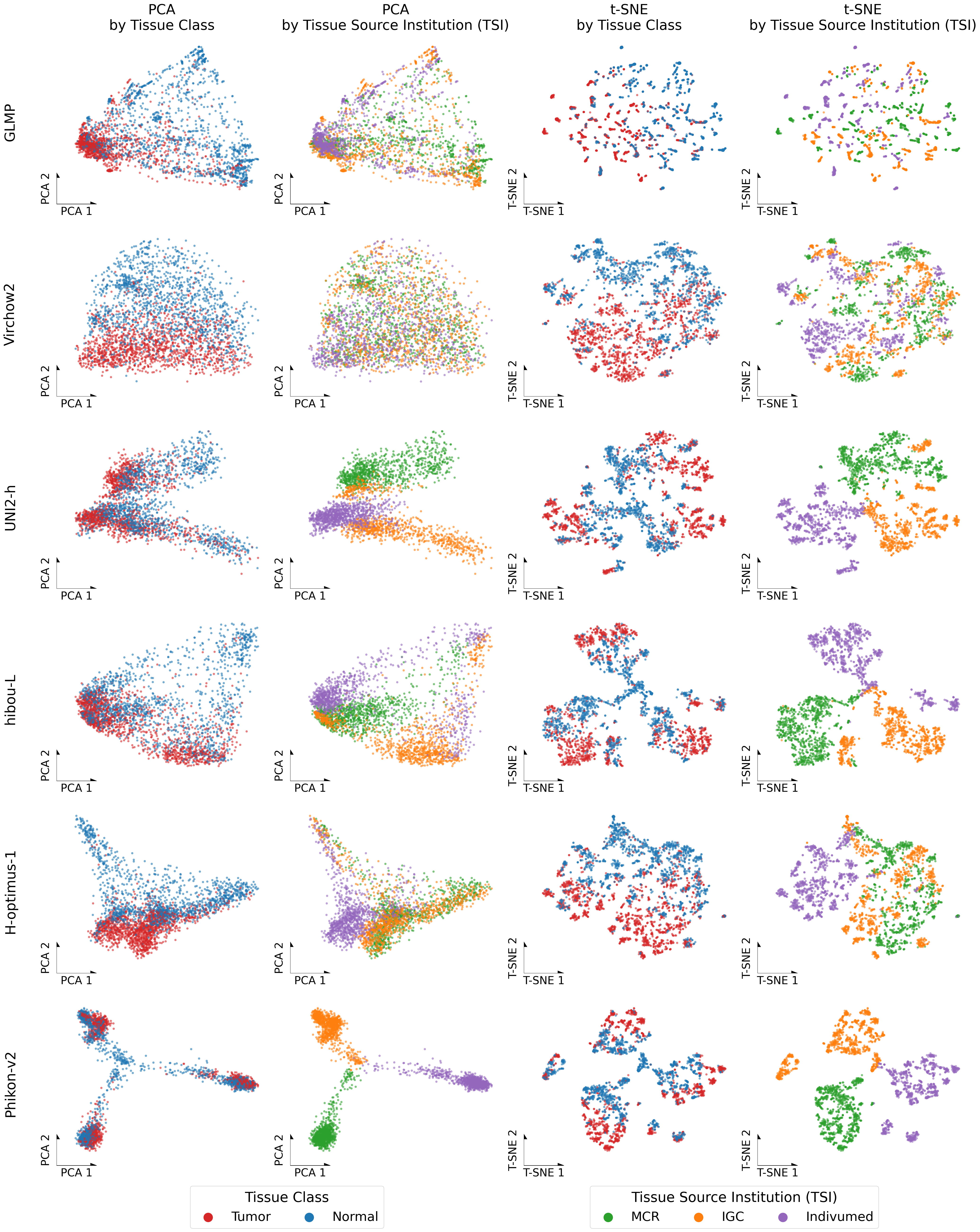}
\caption{Dimension reduction of pathology model embeddings on the TCGA-LUSC Dataset.}
\end{center}
\end{figure}

\begin{figure}[t!]
\ContinuedFloat
\begin{center}
    
    \includegraphics[width=0.9\linewidth]{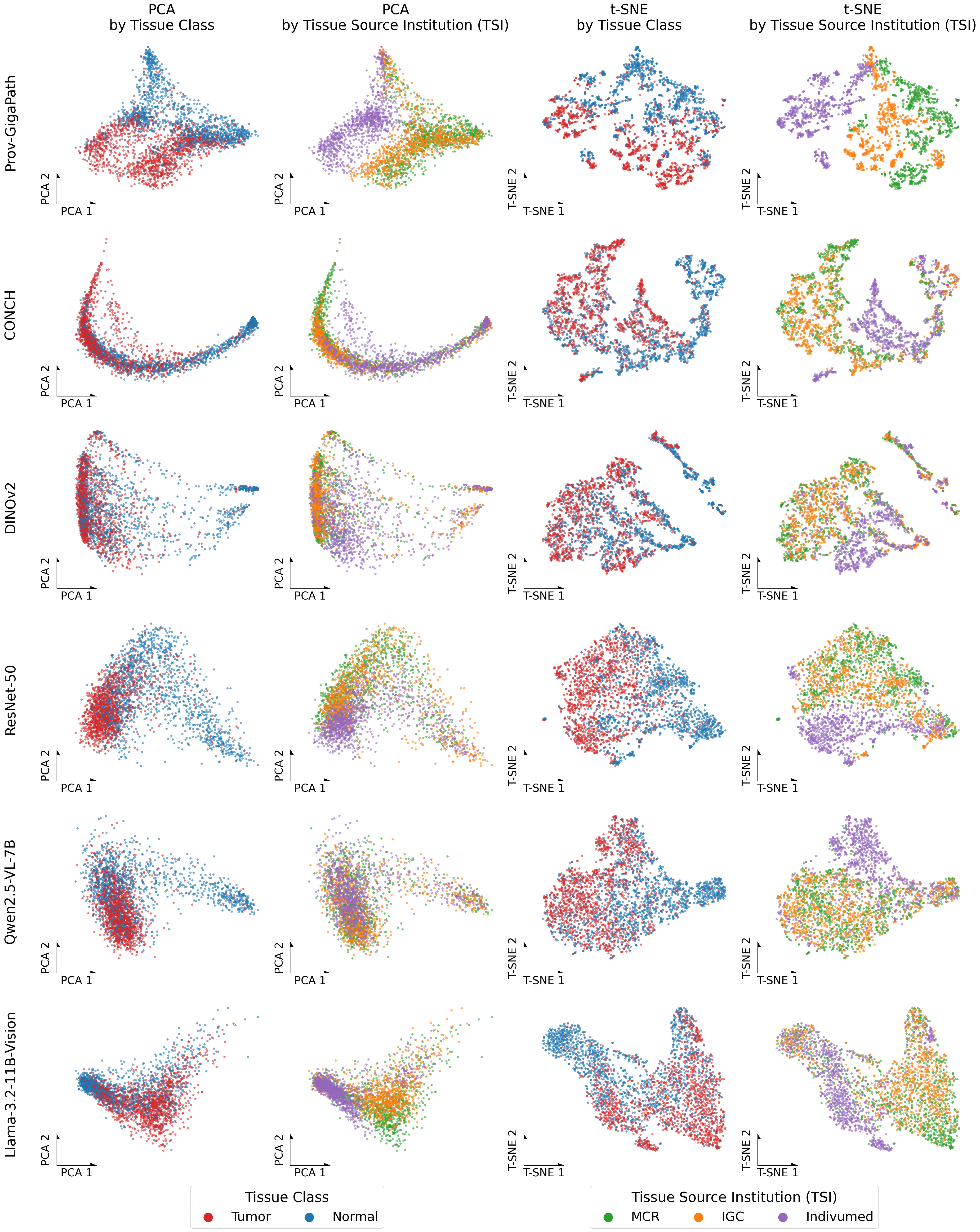}
\end{center}
\caption{(Continued) Dimension reduction of pathology model embeddings on the TCGA-LUSC Dataset.}
\end{figure}

\clearpage
\section{TSI confounding}
\label{sec:tsi confounding}
\subsection{Details for TCGA-LUSC TSI confounding experiments}
\label{sec:tcga tsi confounding}

To evaluate the models on the TCGA-LUSC cohort, we employed a 3-fold cross-validation strategy using samples from three TSIs. In each fold, we grouped two sites together and separated the third to construct spurious correlations between the TSI and the tissue class.

For instance, in Fold 1, TSI-66 was assigned to represent one grouping, while TSI-22 and TSI-56 represented the other. We partitioned the data into four distinct splits with increasing bias levels, sampling exactly 200 patches per WSI. At the 100/0 completely confounded split, all normal training patches originated exclusively from TSI-66, while all tumor training patches originated from TSI-22 and TSI-56. The corresponding test set perfectly reversed this correlation. The precise patch distribution for Fold 1 is shown in \autoref{tab:shortcut_data_tcga}.

\begin{table}[ht]
\centering
\caption{Representative dataset split (Fold 1) for the 3-Fold TSI confounding experiment on TCGA-LUSC. Patch counts are based on sampling 200 patches per WSI. The training sets introduce spurious correlations between the tissue class and TSI, while the testing set features a completely reversed correlation.}
\label{tab:shortcut_data_tcga}
\begin{tabular}{@{}ll rrrr r@{}}
\toprule
& & \multicolumn{2}{c}{\textbf{TSS 66}} & \multicolumn{2}{c}{\textbf{TSS 22 \& TSS 56}} & \\
\cmidrule(lr){3-4} \cmidrule(lr){5-6}
\textbf{Split} & \textbf{Set} &
\textbf{\makecell{\# \\ Normal}} &
\textbf{\makecell{\# \\ Tumor}} &
\textbf{\makecell{\# \\ Normal}} &
\textbf{\makecell{\# \\ Tumor}} &
\textbf{\makecell{Total \\ Patches}} \\
\midrule
50/50 (No Bias)   & Training & 600  & 600 & 600 & 600  & 2,400 \\
67/33 (Low Bias)  & Training & 800  & 400 & 400 & 800  & 2,400 \\
83/17 (Med Bias)  & Training & 1,000& 200 & 200 & 1,000& 2,400 \\
100/0 (High Bias) & Training & 1,200& 0   & 0   & 1,200& 2,400 \\
\midrule
 & Testing     & 0    & 800 & 800 & 0    & 1,600 \\
\bottomrule
\end{tabular}
\end{table}

\clearpage
\subsection{Details for TCGA-LUSC TSI confounding experiments}
\label{Impact of Reinhard Normalization on Shortcut Learning}
\label{app:reinhard}

In the main text, we compare the unnormalized GLMP to baseline PFMs utilizing Macenko stain normalization. To ensure a comprehensive evaluation of stain normalization techniques, we also evaluated the baseline PFMs using Reinhard normalization \cite{reinhard2001color} on CAMELYON16.

\begin{figure}[h!]
\begin{center}
    \includegraphics[width=0.7\textwidth]{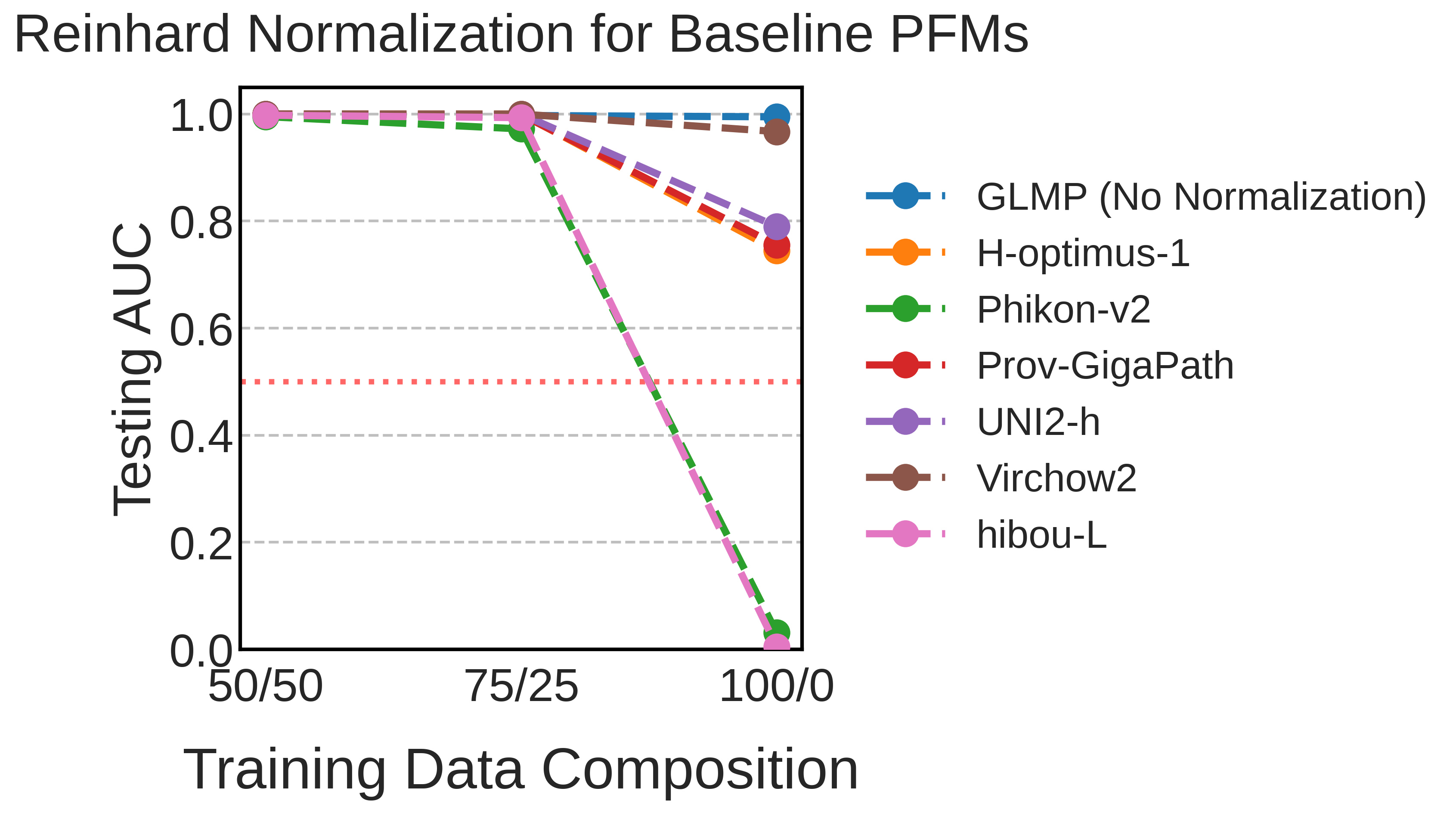}
\end{center}
  \caption{Tissue classification performance under increasing correlations between class label and TSI in CAMELYON16, using Reinhard stain normalization for the baseline PFMs.}
\label{fig:reinhard_shortcut}
\end{figure}

As shown in \autoref{fig:reinhard_shortcut}, Reinhard normalization proved less effective than Macenko normalization at baseline PFMs from shortcut learning, consistent with findings in \cite{komen2025towards}. The performance of several baseline PFMs stagnates near an AUC of zero under the 100/0 high-bias condition. This indicates that the models still heavily relied on residual TSI-specific color statistics that Reinhard normalization failed to remove.

\clearpage
\section{Neighborhood-based Robustness Evaluation}

We evaluated the Robustness Index (RI) on a strictly balanced CAMELYON16 subcohort to quantify the influence of technical artifacts on local embedding geometries. Higher RI values confirm that embedding neighborhoods are driven by true biological similarity rather than site-specific confounders.

\begin{table}[ht]
\centering
\caption{Robustness Index (RI) evaluated on a balanced CAMELYON16 subcohort. Higher values indicate that a patch's local neighborhood in the embedding space is driven by biological similarity rather than confounding TSI-specific artifacts.}
\label{tab:robustness_index}
\begin{tabular}{@{}lc@{}}
\toprule
\textbf{Model} & \textbf{Robustness Index (RI)} \\
\midrule
GLMP (Ours)   & \underline{0.789} \\
Virchow2      & 0.773 \\
H-optimus-1   & 0.697 \\
Prov-GigaPath & 0.405 \\
UNI2-h        & 0.202 \\
hibou-L       & 0.114 \\
Phikon-v2     & 0.024 \\
\bottomrule
\end{tabular}
\end{table}

\clearpage
\section{TSI Prediction}
\label{sec:TSI_pred_table}

\subsection{Linear probe TSI prediction}
\autoref{tab:med_center_pred} expands upon the TSI prediction results presented in the main text by including the standard deviations across 5-fold cross-validation. Accuracies near random chance indicate strong invariance to non-biological batch effects.

\begin{table}[h!]

\centering

\caption{Model susceptibility to batch effects, measured by accuracy on the TSI prediction task using a linear probe. Lower
accuracy indicates less potential influence by TSI-specific signatures and higher robustness to technical artifacts, with a truly TSI-agnostic model expected to perform no better than random chance. Standard deviations were computed across five cross-validation folds using slide-level splits.}
\begin{tabular}{@{}lrrrr@{}}
\toprule
Model & CAMELYON16 & TCGA-LUSC & AI4SkIN & TumSeg \\
\midrule
\makecell[l]{Completely random \\(ideal performance)} & 0.5000 & 0.3333 & 0.5000 & 0.5000 \\
GLMP (ours) & \underline{0.6212 $\pm$ 0.0678} & \underline{0.4750 $\pm$ 0.1053} & \underline{0.5287 $\pm$ 0.0757} & \underline{0.6154 $\pm$ 0.1564} \\
Virchow2 & 0.9998 $\pm$ 0.0002 & 0.9600 $\pm$ 0.0589 & 0.9926 $\pm$ 0.0136
 & 0.9209 $\pm$ 0.0305 \\
UNI2-h & 0.9999 $\pm$ 0.0001 & 0.9690 $\pm$ 0.0440 & 0.9965 $\pm$ 0.0045
 & 0.9201 $\pm$ 0.0365
 \\
hibou-L & 0.9995 $\pm$ 0.0008 & 0.9593 $\pm$ 0.0572 & 0.9983 $\pm$ 0.0016
 & 0.8962 $\pm$ 0.0512 \\
Midnight-12k & 0.9972 $\pm$ 0.0061 & 0.9921 $\pm$ 0.0083 & 0.9700 $\pm$ 0.0425 & 0.8906 $\pm$ 0.0509 \\
H-optimus-1 & 0.9999 $\pm$ 0.0001 & 0.9794 $\pm$ 0.0288 & 0.9914 $\pm$ 0.0058 & 0.8896 $\pm$ 0.0579 \\
Phikon-v2 & 0.9999 $\pm$ 0.0001 & 0.9973 $\pm$ 0.0035 & 0.9894 $\pm$ 0.0156 & 0.9027 $\pm$ 0.0435 \\
Prov-GigaPath & 0.9997 $\pm$ 0.0002 & 0.9560 $\pm$ 0.0498 & 0.9893 $\pm$ 0.0203 & 0.9278 $\pm$ 0.0326 \\
CONCH & 0.9985 $\pm$ 0.0004 & 0.7222 $\pm$ 0.0859 & 0.9536 $\pm$ 0.0205
 & 0.8559 $\pm$ 0.0401 \\
DINOv2-base & 0.9966 $\pm$ 0.0010 & 0.7335 $\pm$ 0.0942 & 0.9640 $\pm$ 0.0258 & 0.8395 $\pm$ 0.0635 \\
ResNet-50 & 0.9877 $\pm$ 0.0098 & 0.8316 $\pm$ 0.0779 & 0.9690 $\pm$ 0.0265 & 0.8030 $\pm$ 0.0832 \\
Qwen2.5-VL-7B-Instruct & 0.9990 $\pm$ 0.0004 & 0.8124 $\pm$ 0.0852 & 0.9907 $\pm$ 0.0063 & 0.8825 $\pm$ 0.0734 \\
Llama-3.2-11B-Vision & 0.9976 $\pm$ 0.0012 & 0.6665 $\pm$ 0.0644 & 0.9813 $\pm$ 0.0140 & 0.8467 $\pm$ 0.0546 \\
\bottomrule
\end{tabular}
\end{table}

\clearpage
\subsection{Details for PC-based TSI prediction experiments}

TSI prediction using top principal components (PCs). A KNN classifier is trained on the top PCs of the embeddings based on a protocol adapted from \cite{komen2024batch}.

\begin{figure}[ht]
\begin{center}
  \includegraphics[width=1\linewidth]{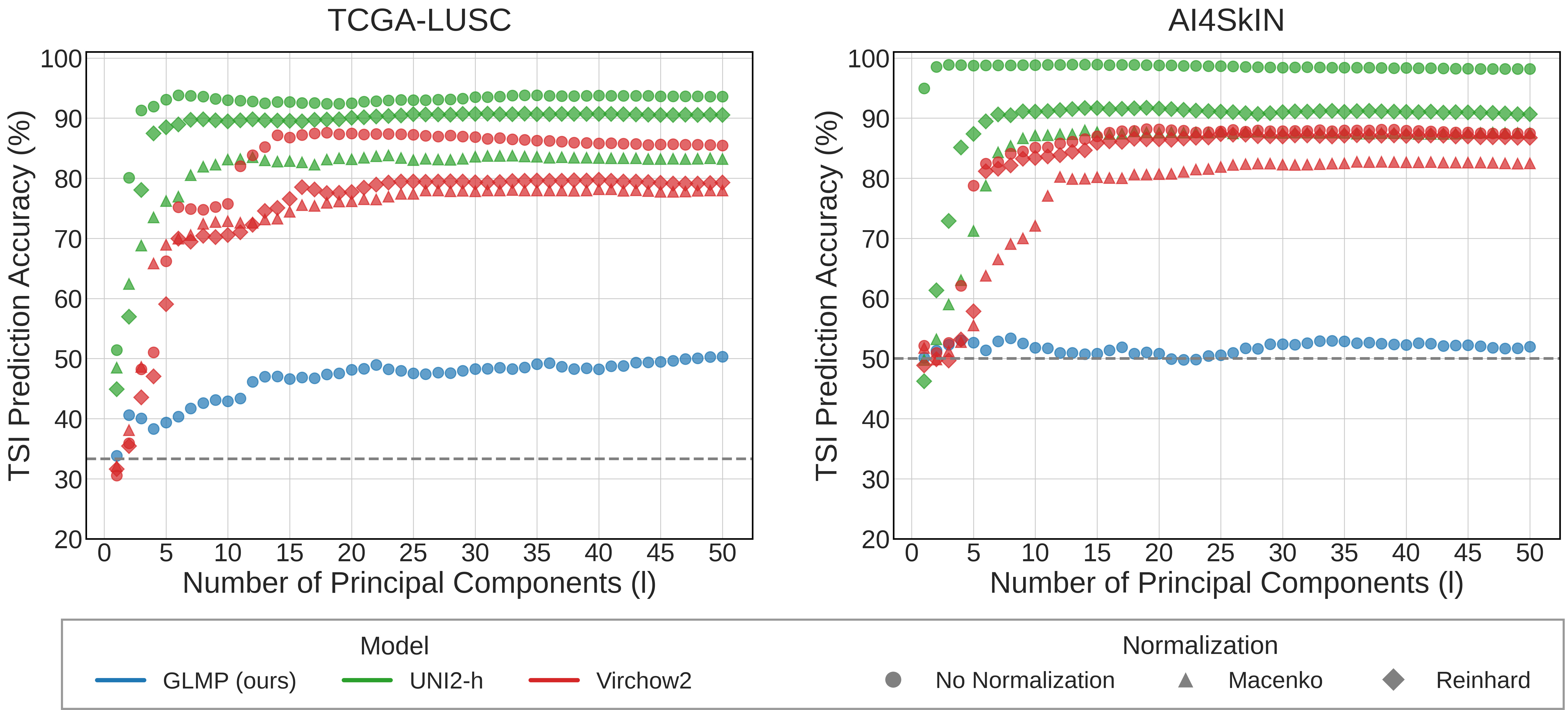}
\end{center}
  \caption{Accuracy for predicting TSI in TCGA-LUSC and AI4SkIN using KNN based on the top principal components (PCs) of the embeddings generated by different models. The horizontal dashed line indicates random chance.}
\label{fig:pc_knn}
\end{figure}

\clearpage
\section{Results of ablation studies}

\subsection{Effect of MLLM prompt focus}

We modified the MLLM prompt to explicitly describe non-biological characteristics (e.g., stain color, scanner focus) to test its filtering capability.
\begin{table}[h!]
\small
\centering
\caption{Effects of MLLM prompt focus on TSI prediction on CAMELYON16. Lower scores indicate better robustness against batch effects.}
\label{tab:prompt_ablation}
\begin{tabular}{@{}lr@{}}
\toprule
MLLM Prompt Focus & TSI prediction accuracy \\
\midrule
Biological features only & 0.62 \\
Biological + non-biological features & 0.73 \\
\bottomrule
\end{tabular}
\end{table}

\subsection{Sensitivity to clustering and text encoders}

To address the sensitivity of the model to the initial grouping and embedding components, we evaluate alternative configurations for the clustering and text encoding models. While our default implementation uses $k$-means ($k=10$) and the Gemini text encoder, we test variations including alternative clustering algorithms (Spectral Clustering \cite{ng2001spectral}, Mini-Batch KMeans \cite{sculley2010web}), varying levels of cluster granularity ($k \in \{8, 12, 15\}$), and alternative text encoders ranging from large general-purpose models (Qwen3 \cite{qwen3embedding}) to smaller domain-specific ones (PubMedBERT \cite{gu2021domain}).

\begin{table}[h!]
    \centering
    \caption{Sensitivity analysis of GLMP components on tissue classification (AUC). We compare the default configuration (KMeans $k=10$, Gemini Embedding) against alternative clustering strategies and text encoders.}
    \label{tab:sensitivity_ablation}
    \vspace{-5pt}
    \footnotesize 
    \setlength{\tabcolsep}{3pt}
    \begin{tabular}{@{}lrrrr@{}}
    \toprule
    & \multicolumn{2}{c}{CAMELYON16} & \multicolumn{2}{c}{TCGA-LUSC} \\
    \cmidrule(lr){2-3} \cmidrule(l){4-5}
    Model Variation & Within-TSI & Cross-TSI & Within-TSI & Cross-TSI \\
    \midrule
    \textbf{GLMP (default)}            & 0.9653 & 0.9714 & 0.8819 & 0.8913 \\
    \midrule
    \multicolumn{5}{l}{\textit{Clustering Strategy}} \\
    KMeans ($k=8$)     & 0.9650 & 0.9706 & 0.8675 & 0.8745 \\
    KMeans ($k=12$)     & 0.9806 & 0.9772 & 0.8858 & 0.8973 \\
    KMeans ($k=15$)     & 0.9748 & 0.9681 & 0.8888 & 0.8902 \\
    Mini-Batch KMeans ($k=10$) & 0.9668 & 0.9641 & 0.8793 & 0.8796 \\
    Spectral Clustering ($k=10$)   & 0.9541 & 0.9486 & 0.8671 & 0.8606 \\
    \midrule
    \multicolumn{5}{l}{\textit{Text Encoder}} \\
    Qwen3-Embedding-8B   & 0.9665 & 0.9707 & 0.8788 & 0.8863 \\
    Pubmedbert-Base-Embeddings  & 0.9589 & 0.9669 & 0.8845 & 0.8853 \\
    \bottomrule
    \end{tabular}
\end{table}

\clearpage
\section{MLLM prompt for GLMP}
\label{sec:prompts}
\subsection{Standard prompt}
\label{sec:standard_prompts}
The following prompt guides the MLLM to generate biology-focused, artifact-free semantic descriptions for histology images, with \_\_ORGAN\_\_ and \_\_SPECIES\_\_ adapted to the specific dataset.
\begin{tcolorbox}[colframe=lightgray, colback=white, boxrule=1pt, arc=2mm, left=0pt, right=0pt, top=0pt, bottom=0pt, boxsep=8pt]

You are a board-certified \_\_ORGAN\_\_ pathologist. 
\vspace{2mm}

\textbf{Input Structure \& Order}

You will receive this entire set of instructions first. Following these instructions, a sequence of content will be provided:
\begin{enumerate}[leftmargin=*, label=\arabic*.]

\item A text line identifying a group, for example: ``The subsequent image patches pertain to Group 1.''
\item Immediately following this text, all image patches belonging to Group 1 will be provided. 
\item This pattern will repeat for subsequent groups (e.g., ``The subsequent image patches pertain to Group 2.'', followed by its images, and so on for all available groups).
\end{enumerate}

Your task is to process all groups and then generate a single JSON output summarizing each one.

\vspace{2mm}
\textbf{Context for Image Analysis}

The H\&E image patches you will receive are from a single whole-slide image (WSI) of a \_\_ORGAN\_\_ tissue from \_\_SPECIES\_\_.

\begin{itemize}[leftmargin=*]
\item \textbf{Group Delineation}: You must strictly use the textual group identifiers provided in the input stream (e.g., ``The subsequent image patches pertain to Group 1.'') to define which images belong to which group, the order of the groups, and for keying your final JSON output.
\item \textbf{Intra-Group Similarity}: All patches within a specific group (as defined by its preceding textual identifier) are expected to share similar histologic features, though minor variations may exist.
\item \textbf{Expected Content}: Depending on the specific tissue type indicated above, you may encounter a wide range of histologic components. The specific features will be pertinent to the organ/tissue system being examined.
\end{itemize}
\vspace{2mm}
\textbf{Tasks}
\begin{enumerate}[leftmargin=*]
\item \textbf{Within-Group Synthesis}: For each group of images presented (e.g., those following ``The subsequent image patches pertain to Group 1.''), carefully examine \textbf{all} provided image patches belonging to that specific group. Synthesize a representative description that captures the \textbf{predominant, consistent, and defining} histologic features observed across these patches. For each group, your synthesis should specifically address the following aspects, which will directly correspond to the fields in the structured summary:
\begin{itemize}[leftmargin=*]
\item \textbf{Architectural Pattern}: Describe the predominant tissue arrangement (e.g., infiltrative growth, preserved native architecture, glandular formation, solid sheets) and structure density. If acellular or non-architectural, describe that arrangement.
\item \textbf{Cellular Morphology \& Cytologic Grade}: Describe predominant cell features: relative cell size, cell density, nuclear pleomorphism, chromatin, nucleoli, mitotic activity, and cytoplasm. Assign an overall nuclear grade (low, intermediate, high) if applicable. If acellular, state `Not applicable'.
\item \textbf{Key Structural Interface}: Describe the most significant structural boundary or architectural interface observed. Focus on the relationship between the main lesion and surrounding tissue, such as the status of a capsule, basement membrane, or the nature of a tumor-stroma interface. If not applicable, state so.
\item \textbf{Stromal Response \& Inflammation}: Describe the stroma and the type, density, and location of any inflammatory infiltrate. If the group is predominantly stroma or inflammation, describe it here.
\item \textbf{Necrosis \& Other Key Features}: Describe the presence and type of any necrosis. Report other key features only if they are diagnostically significant AND truly widespread across the majority of patches, thereby defining the group's overall character. Omit minor, focal, or incidental findings.
\end{itemize}

\item \textbf{Internal Comparison (Mental Step Only - Do Not Output)}:
Mentally compare the synthesized features of each group against the others. This step is for you to refine your within-group synthesis (Task 1). Observing significant differences in key features between groups confirms they are distinct entities. Use this mental differentiation to ensure that the description for each group accurately and uniquely captures \textit{its own} predominant characteristics. \textbf{Absolutely do NOT mention these mental comparisons or reference any other group in your final written output.} Each group's description must be entirely self-contained.
\end{enumerate}
\end{tcolorbox}

\begin{tcolorbox}[colframe=lightgray, colback=white, boxrule=1pt, arc=2mm, left=0pt, right=0pt, top=0pt, bottom=0pt, boxsep=8pt]

\begin{enumerate}[start=3, leftmargin=*]

\item \textbf{Structured Summary (Final Output)}: After all groups and their images have been presented and analyzed, produce a \textbf{single JSON object}. This object will contain one top-level key for each group processed. The keys in the JSON (e.g., ``Group 1'', ``Group 2'') must exactly match the group numbering specified in the textual identifiers that introduced each image set (e.g., ``Group 1'' from ``The subsequent image patches pertain to Group 1.''). Base the descriptions for each field \textbf{strictly} on your synthesis from Task 1 for that specific group.

Example JSON structure:
\begin{tcolorbox}[colframe=gray!60, colback=gray!10, boxrule=1pt, arc=2mm, boxsep=5pt, left=10pt]
\footnotesize
\begin{verbatim}
{
  "Group 1": {
    "Architectural Pattern": "<description>",
    "Cellular Morphology & Cytologic Grade": "<description>",
    "Key Structural Interface": "<description>",
    "Stromal Response & Inflammation": "<description>",
    "Necrosis & Other Key Features": "<description>"
  },
  "Group 2": {
    // ... same structure ...
  }
}
\end{verbatim}
\end{tcolorbox}

\end{enumerate}
\vspace{2mm}
\textbf{Style \& Constraints for Output}
\begin{itemize}[leftmargin=*]
\item \textbf{Focus on Predominant, Defining, and Widespread Features}: Your descriptions MUST reflect features that are predominant, consistent, and diagnostically significant for the entire group of images. Base descriptions on observations from the clear majority of patches. AVOID over-reporting minor or focal variations.
\item \textbf{Absolutely NO Cross-Group References}: Under \textbf{NO circumstances} should the description for one group (e.g., Group 8) mention, compare itself to, or reference any other group. Each group's description MUST be entirely self-contained.
\item \textbf{Do Not Mention Metadata in Descriptions}: Avoid mentioning patch size, number of patches, magnification, or the concept of clustering itself within the descriptive string values of the JSON.
\item \textbf{Conciseness and Terminology}: Keep each field's description to 1-2 sentences. Be precise and use standard histopathology terminology.
\item \textbf{Strictly Avoid Artifact Description}: Focus exclusively on biological features. Omit descriptions of technical artifacts unless they pervasively obscure the biological assessment of a group.
\item \textbf{Strict Output Format}: Output \textbf{only} the JSON object—no extra introductory text, commentary, explanations, or apologies. The output must start directly with `\texttt{\{}' and end directly with `\texttt{\}}'.
\end{itemize}

\vspace{2mm}
The subsequent image patches pertain to Group 1.

\begin{flushleft}
    \includegraphics[width=0.6\linewidth]{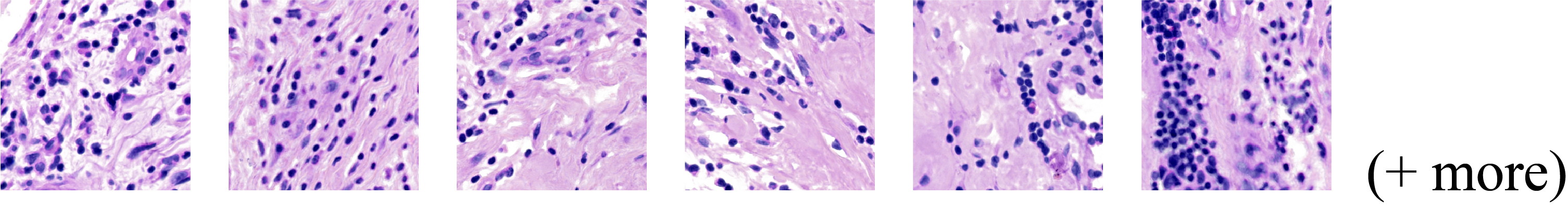}
\end{flushleft}
\vspace{-2mm}
The subsequent image patches pertain to Group 2.

\begin{flushleft}
    \includegraphics[width=0.6\linewidth]{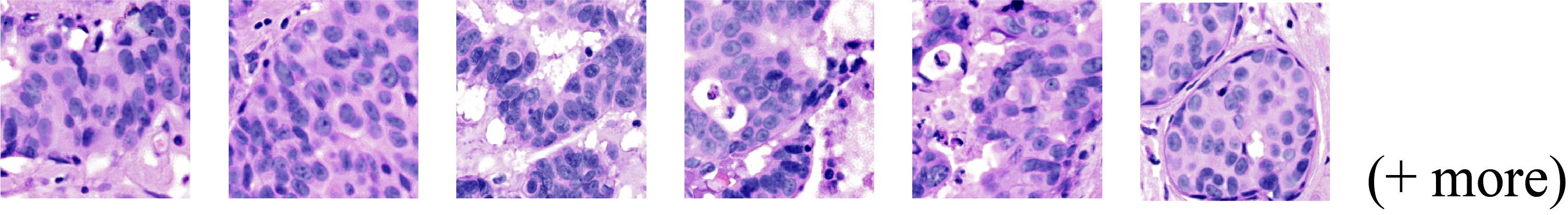}
\end{flushleft} 

\begin{flushleft}
  \hspace{4cm}\textbf{$\vdots$}
\end{flushleft}

The subsequent image patches pertain to Group 10.

\begin{flushleft}
    \includegraphics[width=0.6\linewidth]{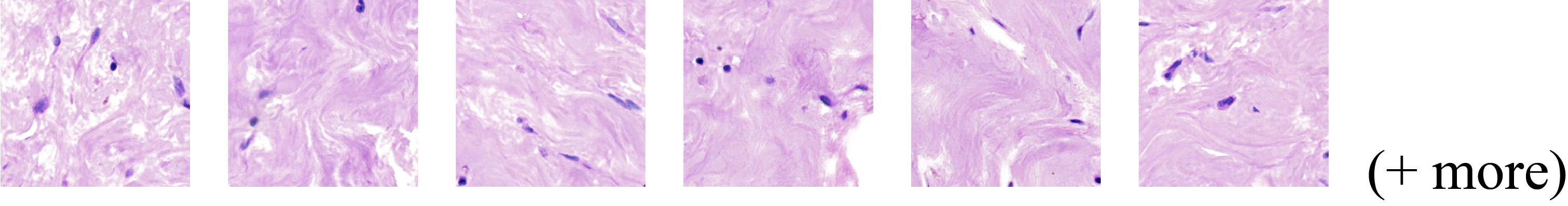}
\end{flushleft} 

\end{tcolorbox}

\clearpage
\subsection{Prompt used in the ablation study}
\label{sec:non_bio_prompts}

This prompt is identical to the biological-only prompt in Appendix~\ref{sec:standard_prompts}, except for the following additions and edits to request descriptions of non-biological artifacts in the ablation study.

\begin{tcolorbox}[colframe=gray!60, colback=white, boxrule=1pt, arc=2mm, left=0pt, right=0pt, top=0pt, bottom=0pt, boxsep=8pt, title=\textbf{Addition to Task 1 fields}]

\begin{itemize}[leftmargin=*]
\item \textbf{Non-Biological Characteristics}: Describe consistent non-biological features that characterize the group, including stain color balance or strength, scanner focus or stitching artifacts, and site or batch cues. Report only if these characteristics are widespread within the group.
\end{itemize}
\end{tcolorbox}

\paragraph{Edits to Style and Constraints:} Replace ``Strictly Avoid Artifact Description'' with: ``Report non-biological characteristics only when they are consistent and pervasive within the group. Do not over-interpret minor or focal artifacts.''

\medskip

\begin{tcolorbox}[colframe=gray!60, colback=gray!10, boxrule=1pt, arc=2mm, boxsep=8pt, left=10pt, title=\textbf{JSON Output }]

\begin{verbatim}
{
  "Group 1": {
    "Architectural Pattern": "<description>",
    "Cellular Morphology & Cytologic Grade": "<description>",
    "Key Structural Interface": "<description>",
    "Stromal Response & Inflammation": "<description>",
    "Necrosis & Other Key Features": "<description>",
    "Non-Biological Characteristics": "<description>"
  },
  "Group 2": {
    // ... same structure ...
  }
}
\end{verbatim}
\end{tcolorbox}

\subsection{Ablation: the MLLM-only approach for tissue classification}
\label{sec:direct_prompts_patch}

\begin{tcolorbox}[colframe=gray!60, colback=white, boxrule=1pt, arc=2mm, left=0pt, right=0pt, top=0pt, bottom=0pt, boxsep=8pt,
title=\textbf{Prompt for the MLLM-only approach for CAMELYON16 tissue classification}]
You are a board-certified lymph node pathologist. Your task is to analyze a single H\&E stained image patch provided to you. This patch is from a larger whole-slide image (WSI) of a lymph node. 
\vspace{2mm}

Your analysis must be strictly confined to the features visible within this single patch. Do not make a diagnosis for the entire slide or the patient. 

\vspace{2mm}

Based on your expert assessment of the morphology, classify the patch. Your response must be a single, valid JSON object and nothing else. The JSON object must contain a single key, ``\texttt{classification}'', and its value must be one of two exact strings: ``\texttt{Tumor}'' or ``\texttt{Normal}''. 

\vspace{2mm}

For example, if you classify the patch as Normal, your \textbf{entire response} must be:

\begin{tcolorbox}[colframe=gray!60, colback=gray!10, boxrule=1pt, boxsep=5pt, left=10pt]

\begin{verbatim}
{
"classification": "Normal"
}
\end{verbatim}
\end{tcolorbox}

\end{tcolorbox}

\clearpage
\subsection{Ablation: the MLLM-only approach for cluster-level tissue classification}
\label{sec:direct_prompts_cluster}

\begin{tcolorbox}[colframe=lightgray, colback=white, boxrule=1pt, arc=2mm, left=0pt, right=0pt, top=0pt, bottom=0pt, boxsep=8pt]
You are a board-certified lymph node pathologist.

\textbf{Input Structure \& Order}
\begin{enumerate}[leftmargin=*]
    \item You will receive this entire set of instructions first.
    \item Following these instructions, a sequence of content will be provided:
    \begin{itemize}
        \item A text line identifying a group (e.g., ``The subsequent image patches pertain to Group 1.'').
        \item Immediately following this text, all image patches belonging to that group will be provided.
        \item This pattern repeats for all subsequent groups.
    \end{itemize}
\end{enumerate}
\textbf{Context for Image Analysis}
The H\&E image patches you will receive are from a single whole-slide image (WSI) of a lymph node.
\begin{itemize}[leftmargin=*]
    \item \textbf{Group Delineation}: Use the textual group identifiers strictly to identify image-group membership.
    \item \textbf{Intra-Group Homogeneity}: All patches within a group represent the same tissue type.
\end{itemize}
\textbf{Tasks}
\begin{enumerate}[leftmargin=*]
    \item \textbf{Diagnostic Assessment}: For each group, examine histologic features (morphology, architecture, nuclear grade) to determine malignancy.
    \begin{itemize}
        \item \textbf{Criteria for ``Tumor''}: Classify as ``Tumor'' if patches show metastatic carcinoma, melanoma, lymphoma, or neoplastic cells (high N:C ratio, pleomorphism, hyperchromasia, abnormal mitoses).
        \item \textbf{Criteria for ``Normal''}: Classify as ``Normal'' if patches show benign components (lymphocytes, germinal centers, sinuses), capsule, adipose tissue, fibrosis, or reactive changes.
    \end{itemize}
    \item \textbf{Structured Summary}: Produce a \textbf{single JSON object} classifying every group.
\end{enumerate}

\textbf{Example JSON structure:}
\begin{tcolorbox}[colframe=gray!60, colback=gray!10, boxrule=1pt, arc=2mm, boxsep=5pt, left=10pt]
\begin{verbatim}
{
  "Group 1": {
    "Classification": "Tumor"
  },
  "Group 2": {
    "Classification": "Normal"
  }
}
\end{verbatim}
\end{tcolorbox}
\textbf{Style \& Constraints for Output}
\begin{itemize}[leftmargin=*]
    \item \textbf{Binary Option Only}: Force a decision between ``Tumor'' or ``Normal''. Do not use terms like ``Suspicious'' or ``Atypical''.
    \item \textbf{No Explanations}: Do not include descriptions, reasoning, or confidence scores.
    \item \textbf{Strict Output Format}: Output \textbf{only} the JSON object. No introductory text or markdown formatting. The output must start with \texttt{\{} and end with \texttt{\}}.
\end{itemize}
\end{tcolorbox}

\clearpage
\section{Examples of MLLM output}
\label{sec:appendix_example_output}

Below are some examples of the structured descriptions generated by the MLLM (Gemini 2.5 Pro) for two histology patch groups from a WSI in MSBCD, using the biologically focused prompt in Appendix~\ref{sec:standard_prompts}.

\begin{tcolorbox}[colframe=lightgray, colback=white, boxrule=1pt, arc=2mm, left=0pt, right=0pt, top=0pt, bottom=0pt, boxsep=8pt]

Input Image (Group 1):
\begin{flushleft}
    \includegraphics[width=0.7\linewidth]{figures/prompt_group1.jpg}
\end{flushleft}

Generated Description (Group 1):
\begin{tcolorbox}[
    colframe=gray!60,
    colback=gray!10,
    boxrule=1pt,
    arc=2mm,
    boxsep=0pt,
    left=10pt
]
\footnotesize\ttfamily 
\textbf{"Group 1":} \{
    \newline\hangindent=1em \textbf{"Architectural Pattern":} "Disorganized, infiltrative pattern of mixed inflammatory cells within a fibrous and edematous stroma. No discrete epithelial or glandular structures are identified.",
    \newline\hangindent=1em \textbf{"Cellular Morphology \& Cytologic Grade":} "The cellular population is composed of small, mature lymphocytes with dark nuclei and scant cytoplasm, intermixed with larger histiocytes showing vesicular nuclei and more abundant cytoplasm. Bland spindle cells are also present. Not applicable for nuclear grade.",
    \newline\hangindent=1em \textbf{"Key Structural Interface":} "The inflammatory infiltrate diffusely permeates the stromal connective tissue without forming a distinct interface with a neoplastic or epithelial component.",
    \newline\hangindent=1em \textbf{"Stromal Response \& Inflammation":} "The stroma is fibrotic and edematous, hosting a prominent, diffuse, mixed chronic inflammatory infiltrate of lymphocytes and histiocytes.",
    \newline\hangindent=1em \textbf{"Necrosis \& Other Key Features":} "No overt necrosis is seen. Scattered apoptotic bodies and cellular debris are present within the inflammatory milieu."
\newline\}
\end{tcolorbox}

\vspace{2mm}

Input Image (Group 2):
\begin{flushleft}
    \includegraphics[width=0.7\linewidth]{figures/prompt_group2.jpg}
\end{flushleft} 

Generated Description (Group 2):
\begin{tcolorbox}[
    colframe=gray!60,
    colback=gray!10,
    boxrule=1pt,
    arc=2mm,
    boxsep=0pt, 
    left=10pt,
]
\footnotesize\ttfamily 
\textbf{"Group 2":} \{
    \newline\hangindent=1em \textbf{"Architectural Pattern":} "Cohesive nests of neoplastic epithelial cells arranged in solid and cribriform patterns, consistent with ductal carcinoma in situ. The nests fill and expand duct-like spaces.",
    \newline\hangindent=1em \textbf{"Cellular Morphology \& Cytologic Grade":} "The neoplastic cells exhibit moderate to marked pleomorphism, with enlarged, hyperchromatic to vesicular nuclei, irregular nuclear contours, and prominent nucleoli. The overall features correspond to an intermediate to high nuclear grade (Grade 2-3).",
    \newline\hangindent=1em \textbf{"Key Structural Interface":} "The neoplastic epithelial nests are sharply demarcated from the surrounding stroma, suggesting confinement within a ductal-lobular unit.",
    \newline\hangindent=1em \textbf{"Stromal Response \& Inflammation":} "The periductal stroma is fibrous and contains a mild, patchy lymphoplasmacytic infiltrate.",
    \newline\hangindent=1em \textbf{"Necrosis \& Other Key Features":} "Focal luminal necrosis is present, characterized by eosinophilic, granular debris. Apoptotic cells are also noted within the epithelial nests."
\newline\}
\end{tcolorbox}
\end{tcolorbox}
\end{document}